\documentclass{article}

\usepackage[main, final]{neurips_2026}

\usepackage[utf8]{inputenc} 
\usepackage[T1]{fontenc}    
\usepackage[hypertexnames=false]{hyperref}       
\usepackage{url}            
\usepackage{booktabs}       
\usepackage{multirow}       
\usepackage{amsmath}        
\usepackage{amsfonts}       
\usepackage{amssymb}        
\usepackage{amsthm}         
\usepackage{nicefrac}       
\usepackage{microtype}      
\usepackage{xcolor}         
\usepackage{xspace}         
\usepackage{pifont}         
\usepackage[table]{colortbl} 
\usepackage{graphicx}       
\usepackage{float}          
\usepackage{wrapfig}        
\usepackage{tikz}           
\usetikzlibrary{positioning, arrows.meta, fit, calc, patterns}
\usepackage{titletoc}       

\titlecontents{lsection}[0em]              
  {\addvspace{0.8em}\bfseries}             
  {\thecontentslabel\hspace{0.6em}}        
  {}                                       
  {\hfill\contentspage}                    
\titlecontents{lsubsection}[1.8em]         
  {\addvspace{0.15em}}                     
  {\thecontentslabel\hspace{0.5em}}        
  {}                                       
  {\titlerule*[0.5pc]{.}\contentspage}     
\titlecontents{lsubsubsection}[4.2em]      
  {\addvspace{0.1em}}                      
  {\thecontentslabel\hspace{0.5em}}        
  {}                                       
  {\titlerule*[0.5pc]{.}\contentspage}     

\newif\ifdraft
\draftfalse
\ifdraft
  \newcommand{\noctis}[1]{\textcolor{red}{{\sf (Noctis:} {\sl{#1})}}}
  \newcommand{\xxu}[1]{\textcolor{blue}{{\sf (Xu:} {\sl{#1})}}}
  \newcommand{\todos}[1]{\textcolor{orange}{{\sf (TO-DO:} {\sl{#1})}}}
\else
  \newcommand{\noctis}[1]{}
  \newcommand{\xxu}[1]{}
  \newcommand{\todos}[1]{}
\fi

\theoremstyle{plain}
\newtheorem{proposition}{Proposition}
\newtheorem{corollary}[proposition]{Corollary}
\newtheorem{claim}{Claim}
\theoremstyle{remark}
\newtheorem{remark}{Remark}

\AddToHook{env/wraptable/begin}{%
  \setlength{\abovecaptionskip}{0pt}%
  \setlength{\belowcaptionskip}{7pt}}

\newcommand{\method}{CtrlFlow\xspace}

\newcommand{\cmark}{{\color{green!60!black}\ding{51}}}
\newcommand{\xmark}{{\color{red!60}\ding{55}}}
\newcommand{\pmark}{{\color{orange!80!black}$\boldsymbol{\sim}$}}

\title{Decompose the Distillation: Interpretable Single-Pass Guidance for Diffusion Models}

\author{%
  Wuyang Zhang \qquad Xiaolin Xu\thanks{Corresponding author.} \\
  Department of Electrical and Computer Engineering \\
  Northeastern University, Boston, MA, USA \\
  \texttt{doggo@ieee.org}, \texttt{x.xu@northeastern.edu} \\
}

\begin{document}

\maketitle

\begin{abstract}
Classifier-Free Guidance (CFG) is a widely used inference-time guidance rule for conditional image generation with diffusion models, which exposes a guidance scale that adjusts conditioning strength after training. This flexibility requires conditional and unconditional model evaluations at every sampling step, so an $N$-step trajectory takes $2N$ forward passes. Single-pass alternatives reduce such cost, yet most collapse CFG's two components into one opaque guided prediction, losing the separable base flow and guidance direction that support per-step control and diagnostics, even when the scale remains adjustable.
This work introduces \method, a single-pass guidance method that preserves CFG's separable base-flow and guidance-direction interface.
To preserve both components at one-pass cost, \method attaches two lightweight LoRA adapters to a frozen backbone, each adding $2.5\%$ parameters, and predicts the base flow and guidance direction from one pass of the shared frozen backbone. Because the guidance scale multiplies the guidance direction only after the forward pass, one evaluation yields the guided velocity at any scale, so a single trained checkpoint remains adjustable at inference without retraining. Preserving two outputs introduces a training mismatch: inference amplifies guidance-direction errors by the chosen scale, while naive component-wise losses weight both outputs equally.
To align training with this composed inference-time error, \method uses a unified multi-scale distillation loss.
Residual guidance errors can still persist at extreme scales, where even small guidance-direction errors are amplified by the guidance scale. To reduce this inference-time error, \method applies scale-adaptive modulation to damp velocity-field errors and time-grid warping to concentrate solver evaluations where the trajectory changes most rapidly.
On ImageNet $256\times256$ with the DiT-XL/2 backbone, \method matches CFG within a pre-specified tolerance in 24 of 25 $(N, s)$ configurations, against 20 of 25 for a matched scale-conditioned opaque student, and achieves a measured $1.93\times$ wall-clock speedup at $N{=}25$.
\end{abstract}

\section{Introduction}
\label{sec:introduction}

In conditional image generation with diffusion models~\citep{ho2020ddpm, song2021scorebased, rombach2022ldm}, each sample should both look like a plausible natural image and match the conditioning signal, such as the requested class in class-conditional generation~\citep{dhariwal2021beatgans, peebles2023dit}.
Achieving both goals simultaneously is challenging: stronger guidance improves condition matching, whereas overly strong guidance can introduce visible artifacts~\citep{sadat2024apg} and make the model produce similar outputs instead of diverse valid images for the same condition~\citep{karras2024autoguidance, kynkaanniemi2024interval}.
Inference-time guidance methods~\citep{dhariwal2021beatgans, ho2022cfg} address this trade-off by letting practitioners control the strength of the conditioning signal during sampling.
Classifier-Free Guidance (CFG)~\citep{ho2022cfg}, the most widely adopted such method and a foundation for many later variants, implements this control with a scalar guidance scale $s$ that can be tuned after training.
Because CFG compares two predictions from the same network at every sampling step, practitioners obtain per-sample, per-step control over the quality-alignment trade-off without retraining.
This flexibility carries a cost: every guided step requires two network evaluations rather than one, so an $N$-step trajectory takes $2N$ forward passes and wall-clock latency nearly doubles~\citep{ho2020ddpm, song2021scorebased, rombach2022ldm}.

Several recent works attacked this $2N$ overhead.
Across these attempts, speed is typically gained by giving up some facet of CFG's interface, leaving practitioners to choose between latency and control.
One route distills guidance into a single-pass student that takes the scale as a network input, as in AGD~\citep{sadat2025agd}, which trains an adapter on a frozen backbone, and Guidance-Free Training (GFT)~\citep{chen2025gft}, which fine-tunes the whole model.
Both halve the forward passes and keep the scale adjustable, but each pass commits to one scale and returns an opaque guided velocity with no separable base flow or guidance direction.
Model Guidance~\citep{tang2025modelguidance} and Entropy Rectifying Guidance~\citep{berrada2025erg} take a different route, building guidance directly into the model.
Both methods run in a single pass, yet both collapse the base/guidance split and surrender the post-hoc scale knob that motivated CFG.
A complementary pattern keeps two passes and improves the guidance step instead.
CFG-Zero*~\citep{fan2025cfgzero} and APG~\citep{sadat2024apg} sharpen the guided update, while retaining the conditional and unconditional evaluations.
Autoguidance~\citep{karras2024autoguidance} makes the second evaluation cheaper by replacing the unconditional model with a smaller trained network.
None of these two-pass refinements reduces the $2N$ evaluation count.
For the goal of CFG-like control at lower latency, these trade-offs reduce to three axes.
\emph{Single-pass compute} measures whether each step uses one network evaluation or two.
The criterion of \emph{explicit access to $\mathbf{v}_0$ and $\mathbf{g}$} concerns whether the method's inference interface keeps base flow and guidance direction separately available for control or diagnosis.
The remaining axis, \emph{post-hoc $s$}, captures whether the method exposes a guidance-scale knob that is not tied to a single training scale and can be varied at inference without retraining.
Table~\ref{tab:method_comparison} organizes existing methods along these axes.
Scale-conditioned students such as AGD and GFT already combine single-pass compute with post-hoc $s$, so the missing piece is the interface: to our knowledge, no prior single-pass method keeps $\mathbf{v}_0$ and $\mathbf{g}$ separately available.

\begin{wraptable}[23]{r}{0.48\textwidth}
  \centering
  \vspace{-1.8em}
  \caption{Comparison of guidance methods. \emph{Explicit $\mathbf{v}_0/\mathbf{g}$}: base flow and guidance direction separately accessible. \emph{Post-hoc $s$}: guidance scale adjustable without retraining.}
  \label{tab:method_comparison}
  \scriptsize
  \setlength{\tabcolsep}{3pt}
  \begin{tabular}{@{}l ccc@{}}
    \toprule
    \textbf{Method} & \shortstack{\textbf{Single}\\\textbf{Pass}} & \shortstack{\textbf{Explicit}\\\textbf{$\mathbf{v}_0/\mathbf{g}$}} & \shortstack{\textbf{Post-hoc}\\\textbf{$s$}} \\
    \midrule
    \multicolumn{4}{l}{\textit{Two-Pass (NFE\,$=$\,2N)}} \\[2pt]
    \quad CFG & \xmark & \cmark\textsuperscript{$\dagger$} & \cmark \\
    \quad CFG + Interval & \xmark & \cmark\textsuperscript{$\dagger$} & \cmark \\
    \quad CFG-Zero* & \xmark & \xmark\textsuperscript{$\S$} & \cmark \\
    \quad APG & \xmark & \pmark\textsuperscript{$\ddagger$} & \cmark \\
    \quad Autoguidance & \xmark & \xmark\textsuperscript{$\P$} & \cmark \\
    \midrule
    \multicolumn{4}{l}{\textit{Single-Pass (NFE\,$=$\,N)}} \\[2pt]
    \quad AGD & \cmark & \xmark & \cmark\textsuperscript{$\|$} \\
    \quad GFT & \cmark & \xmark & \cmark\textsuperscript{$\|$} \\
    \quad Model Guidance & \cmark & \xmark & \xmark \\
    \quad ERG & \cmark & \xmark & \xmark \\
    \rowcolor{blue!6}[0pt][0pt]
    \quad \textbf{\method (Ours)} & \cmark & \cmark & \cmark \\
    \bottomrule
  \end{tabular}\\[0.3em]
  {\footnotesize
  \textsuperscript{$\dagger$}Via two evaluations, not one pass.\\
  \textsuperscript{$\ddagger$}Projects relative to $\mathbf{v}_{\text{cond}}$, two passes.\\
  \textsuperscript{$\S$}Rescales combined update, split not preserved.\\
  \textsuperscript{$\P$}Smaller unconditional net, split not preserved.\\
  \textsuperscript{$\|$}Scale is a network input, one opaque output per pass.}
\end{wraptable}
\vspace{-0.5em}

Motivated by this missing intersection, we observe that CFG's two-prediction structure exposes an interpretable interface between a base flow $\mathbf{v}_0$ and a guidance direction $\mathbf{g}$, which should be preserved rather than collapsed into a single opaque prediction.
The base flow captures how the model moves toward the data manifold without conditioning, while the guidance direction encodes how the conditioning signal redirects that flow.
Existing single-pass efficient methods compress this interface into a single opaque output.
To preserve it at one-pass cost, \method attaches two lightweight LoRA adapters~\citep{hu2022lora}, each adding $2.5\%$ parameters, to a frozen backbone and computes both components in a single forward pass.
As a result, the guidance scale $s$ multiplies the guidance direction after the forward pass, so $s$ remains adjustable at inference and no retraining is needed.
Unlike a scale input, this lets one evaluation yield the guided velocity at every $s$.
In addition to post-hoc scale control, the preserved interface opens a diagnostic channel unavailable to opaque single-pass methods.
Norm maps reveal \emph{where} guidance acts spatially, while alignment maps reveal \emph{whether} guidance reinforces or opposes the base flow, and this geometric signal later enables the alignment-based correction in Section~\ref{subsec:guidance_modulation}.

Preserving the decomposition is not free, however, because distilling this interface exposes three challenges that require explicit handling.

\textbf{(i) Training-pressure mismatch.}
Supervising the two components independently creates a mismatch between training pressure and inference-time error importance within the fixed adapter budget.
Inference weights guidance errors by $s$, but a naive component-wise loss treats both components equally, so the guidance adapter receives weaker gradient signal than its inference-time importance warrants.
A unified distillation objective sidesteps this mismatch by supervising only the composed output, which aligns gradient pressure with the composed inference-time error (Section~\ref{subsec:training}).

\textbf{(ii) Error amplification.}
At high guidance scales with low step counts, small prediction errors in the guidance direction are magnified by $s$, producing visible artifacts.
Scale-adaptive modulation addresses this amplification by targeting three sources of error (Section~\ref{subsec:guidance_modulation}).
Temporal windowing narrows the guidance-active interval as $s$ grows, while scale dampening reduces the guidance magnitude above a tuned threshold.
Alignment-based correction then uses the geometric agreement between base flow and guidance direction, a signal that no opaque single-pass student exposes.

\textbf{(iii) Dead zones.}
In our diagnostics, certain $(N, s)$ configurations show large FID degradation that is better explained by temporal under-resolution of the solver than by a globally inaccurate learned velocity field.
A scale-dependent time-warp controller redistributes the existing $N$ evaluations near critical trajectory regions without adding forward passes (Section~\ref{subsec:timewarp}).
With this warp, all observed dead-zone failures except the extreme $(2, 8)$ setting are resolved (Appendix~\ref{app:deadzone}), leaving 24 of 25 configurations within tolerance.

We evaluate on ImageNet $256\times256$ with the DiT-XL/2 backbone~\citep{peebles2023dit}.
Under the FID tolerance defined in Section~\ref{sec:experiments}, these mechanisms recover CFG-match quality while retaining one-pass sampling.
\method matches CFG quality across all five tested step counts at $s{=}1$ and across four of five at $s{=}8$. At $N{=}25$, wall-clock time falls from $843$\,ms per image to $437$\,ms, a $1.93\times$ speedup.
Beyond the training range of $s \le 8$, \method remains competitive: it outperforms CFG at $s{=}10$ (FID $28.41$ versus $31.24$) and trails CFG by 3.1 FID at the more extreme $s{=}12$ setting in our DiT-XL/2 protocol.
Against a scale-conditioned opaque student with the same capacity, teacher, and curriculum (AGD-MultiScale), \method is on par at $(25, 2)$ and 0.96 to 4.14 FID lower at the harder tested configurations, matches CFG in 24 of 25 configurations against 20 of 25, and keeps top-1 class accuracy within 2 points of CFG.

Our contributions are organized around preserving and exploiting this interface.

$\bullet$~\textbf{Interpretable single-pass guidance interface.}
\method exposes both $\mathbf{v}_0$ and $\mathbf{g}$ in one forward pass with $5\%$ parameter overhead, while keeping the guidance scale $s$ adjustable at inference without retraining.
The dual-adapter architecture and unified distillation objective that enable this interface are described in Sections~\ref{subsec:architecture} and~\ref{subsec:training}.

$\bullet$~\textbf{Decomposition-enabled stabilization.}
\method combines temporal windowing and scale dampening with an alignment-based correction that only the preserved split makes possible, reducing error amplification across the tested range of $s$.
At $(4, 8)$, these modulations lower FID from 73.41 to 63.42 before time warping is applied (Section~\ref{subsec:guidance_modulation}, Table~\ref{tab:contribution_isolation}).
These mechanisms also help opaque students, but with every transferable mechanism given to a matched opaque student, \method remains 1.3 to 2.1 FID lower (Section~\ref{subsec:ablations}).

$\bullet$~\textbf{Dead-zone removal.}
Scale-dependent time-grid warping resolves all observed dead-zone failures except $(2, 8)$, leaving 24 of 25 configurations within tolerance without additional forward passes (Section~\ref{subsec:timewarp}).

$\bullet$~\textbf{CFG-match quality at one-pass cost.}
Across the 25-point $(N, s)$ grid on DiT-XL/2, \method matches CFG quality in 24 of 25 cases, against 20 of 25 for the matched scale-conditioned opaque student, and the measured $N{=}25$ latency shows a $1.93\times$ wall-clock speedup over CFG (Section~\ref{sec:experiments}, Table~\ref{tab:main_results}).
On DiT-L/4, \method stays within 0.19 FID of CFG at $(25, 2)$ and outperforms CFG at the tested high-scale settings, and a Stable Diffusion 2.1 pilot adds text conditioning (Section~\ref{subsec:cross_backbone}).

\section{Background and Related Works}
\label{sec:background}

\textbf{CFG and its Interface.}
Because \method uses a DiT velocity-prediction backbone~\citep{peebles2023dit}, we write diffusion sampling in the flow-matching form~\citep{lipman2023flow}, where a learned velocity field transports noise to data along an ODE.
Classifier-Free Guidance (CFG)~\citep{ho2022cfg} steers conditional generation by interpolating unconditional and conditional predictions,
\begin{equation}
  \mathbf{v}_{\text{cfg}} = \mathbf{v}_{\text{uncond}} + s \cdot \left(\mathbf{v}_{\text{cond}} - \mathbf{v}_{\text{uncond}}\right),
  \label{eq:cfg}
\end{equation}
where $s$ controls guidance strength.
Equivalently, CFG computes $\mathbf{v}_0 + s\,\mathbf{g}$, where $\mathbf{v}_0 \triangleq \mathbf{v}_{\text{uncond}}$ is the base flow and $\mathbf{g} \triangleq \mathbf{v}_{\text{cond}} - \mathbf{v}_{\text{uncond}}$ is the guidance direction.
This decomposition reveals an interpretable interface between unconditional generation and conditioning adjustment, but CFG obtains both components through two evaluations per step, doubling the forward-pass count.

\textbf{Efficient Guidance.}
Existing efficiency methods mostly split between lowering the number of evaluations and preserving CFG's separate outputs.
Single-pass methods such as Plug-and-Play Distillation~\citep{hsiao2024plug}, AGD~\citep{sadat2025agd}, Guidance-Free Training (GFT)~\citep{chen2025gft}, Model Guidance~\citep{tang2025modelguidance}, and ERG~\citep{berrada2025erg} reduce per-step cost by avoiding a separate unconditional evaluation, often through a compressed guide or training-time guidance surrogate.
AGD and GFT take the guidance strength as a network input, so one checkpoint covers many scales, but each pass still returns one opaque guided velocity at one scale.
Two-pass refinements such as APG~\citep{sadat2024apg}, CFG-Zero*~\citep{fan2025cfgzero}, Autoguidance~\citep{karras2024autoguidance}, Dynamic Negative Guidance~\citep{koulischer2025dng}, and ICG~\citep{sadat2025icg} preserve or improve CFG quality, but they retain the $2N$ evaluation cost.
Thus, prior work either preserves the CFG interface at two-pass cost or reduces the pass count by exposing only a single opaque guided output.

\textbf{Schedules and Training Pressure.}
This interface gap also affects distillation and sampling design.
Distillation methods~\citep{salimans2022progressive, meng2023distillation, song2023consistency, luo2023lcm} collapse the two components into a single composed output, removing the separate guidance direction that \method needs for post-hoc control and diagnostics.
Guidance-window and solver-schedule methods~\citep{kynkaanniemi2024interval, malarz2025betacfg, lu2022dpmsolver, sabour2024align, chen2024adaptive} instead modify when or where guidance is applied, but they do not adapt an exposed guidance direction across scales.
Once a student is asked to output both $\mathbf{v}_0$ and $\mathbf{g}$, it also faces a training-pressure mismatch: a naive dual loss treats the two adapter errors equally even though inference amplifies guidance errors by $s$.
\method responds by preserving the $\mathbf{v}_0/\mathbf{g}$ interface in one pass, aligning distillation with the composed inference-time error, and using scale-aware modulation and time-grid warping to adapt that interface across guidance scales.

Section~\ref{sec:methodology} defines the decomposition formally, implements it with dual adapters, trains it with unified distillation, and introduces the inference-time mechanisms that stabilize it across scales and step counts.
Appendix~\ref{app:extended_related} provides an extended discussion of all related work.

\vspace{-0.5em}
\section{Methodology}
\label{sec:methodology}
\vspace{-0.5em}

\begin{figure}[t]
  \centering
  \resizebox{0.67\textwidth}{!}{%
    \includegraphics{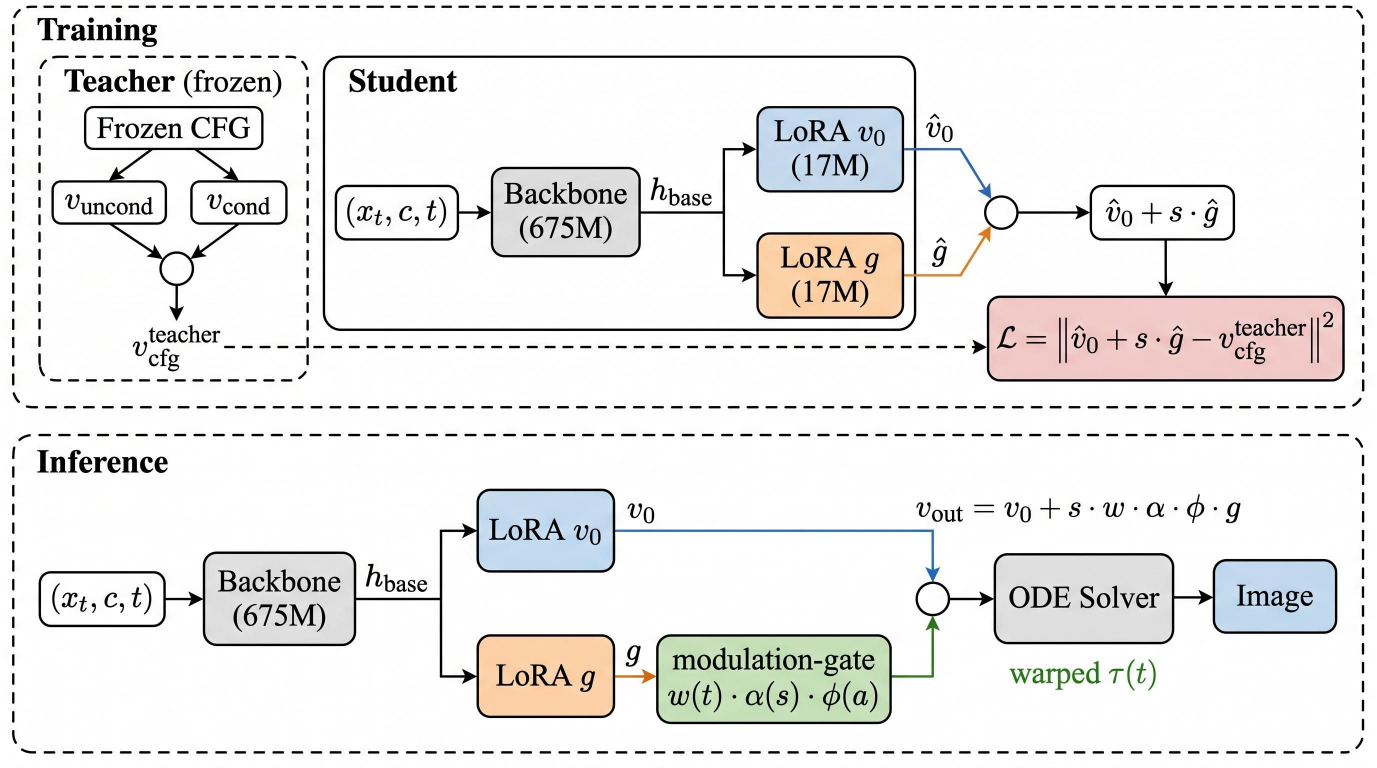}%
  }%
    \vspace{-0.5em}
  \caption{\method compresses CFG to one pass while preserving separate $\mathbf{v}_0$ and $\mathbf{g}$ outputs. \textbf{Top.} A frozen CFG teacher provides the target $\mathbf{v}_{\text{cfg}}^{\text{teacher}}$. The student matches it via a unified loss on $\hat{\mathbf{v}}_0 + s \cdot \hat{\mathbf{g}}$. \textbf{Bottom.} At inference, one backbone pass produces $\mathbf{v}_0$ and $\mathbf{g}$ through two LoRA adapters. The scale $s$ multiplies $\mathbf{g}$ post-hoc. Three modulation gates ($w$, $\alpha$, $\varphi$) stabilize guidance, and a warped time grid concentrates evaluations near high-curvature regions.}
  \vspace{-1.5em}
  \label{fig:method_overview}
\end{figure}

\method distills CFG into a single-pass student that preserves separate base-flow and guidance-direction outputs rather than emitting one opaque guided velocity.
Figure~\ref{fig:method_overview} summarizes this design.
The section first formalizes this output interface and implements it with dual adapters (Sections~\ref{subsec:decomposition} and~\ref{subsec:architecture}).
Because the resulting two-output student can still misallocate its limited adapter capacity (the training-pressure mismatch from Section~\ref{sec:introduction}), we train it with unified multi-scale distillation (Section~\ref{subsec:training}).
Finally, two inference-time mechanisms stabilize sampling: scale-adaptive modulation, whose alignment gate reuses the preserved interface, damps high-scale guidance errors, and time-grid warping adapts the solver schedule to the guidance regime (Sections~\ref{subsec:guidance_modulation} and~\ref{subsec:timewarp}).
\vspace{-0.75em}

\subsection{Velocity Decomposition}
\label{subsec:decomposition}

As shown in Section~\ref{sec:background}, CFG can be written as $\mathbf{v}_0 + s\,\mathbf{g}$, where $\mathbf{v}_0$ is the base flow and $\mathbf{g}$ is the guidance direction.
We now formalize this split as the output interface that the compressed student must preserve.
Given a noisy sample $\mathbf{x}_t$, conditioning signal $c$, and timestep $t$, the two teacher-side fields are
\begin{align}
  \mathbf{v}_0(\mathbf{x}_t, t) &\triangleq \mathbf{v}_\theta(\mathbf{x}_t, \varnothing, t), \label{eq:v0_def} \\
  \mathbf{g}(\mathbf{x}_t, c, t) &\triangleq \mathbf{v}_\theta(\mathbf{x}_t, c, t) - \mathbf{v}_\theta(\mathbf{x}_t, \varnothing, t), \label{eq:g_def}
\end{align}
so the teacher's guided velocity at any scale $s$ is
\begin{equation}
  \mathbf{v} = \mathbf{v}_0 + s \cdot \mathbf{g},
  \label{eq:decomposition}
\end{equation}
with the linear dependence on $s$ explicit.
Although this decomposition is algebraically identical to standard CFG, treating the pair $(\mathbf{v}_0, \mathbf{g})$ as the student's required output interface is what enables the mechanisms that follow: the guidance scale $s$ can be applied after the forward pass, the norm $\|\mathbf{g}\|$ can be inspected spatially, and the alignment between $\mathbf{v}_0$ and $\mathbf{g}$ can drive inference-time gating.
In the compressed student, $\hat{\mathbf{v}}_0$ and $\hat{\mathbf{g}}$ (hatted to distinguish student predictions from teacher-defined fields) are learned estimates of these teacher fields produced from a single shared forward pass that receives $(\mathbf{x}_t, c, t)$. The unconditional teacher evaluation that defines $\mathbf{v}_0$ is used only for supervision and is not repeated at inference.
Below, hats appear when emphasizing training-time predictions. After training, we write the deployed student outputs as $\mathbf{v}_0$ and $\mathbf{g}$ when no confusion with teacher fields arises.

\textbf{Interpretable Diagnostics.}
Because the student preserves both fields as separate outputs, the decomposition exposes two quantities unavailable to single-output methods.
The norm $\|\mathbf{g}\|$ measures the magnitude of the conditioning-induced adjustment at each spatial location and timestep: large values indicate strong redirection, while small values mark effectively unconditional regions.
The alignment $a \triangleq \cos(\mathbf{v}_0, \mathbf{g})$ quantifies compatibility between the base flow and the conditioning adjustment.
When $a \approx 1$, conditioning reinforces the base flow, and when $a < 0$, conditioning opposes it, producing a diagnostic signal for potential artifacts.
Section~\ref{subsec:guidance_modulation} uses $a$ as an inference-time modulation signal, and Section~\ref{subsec:interpretability} visualizes both quantities on generated samples.
These diagnostics are only useful if compression keeps both fields accessible, which the next subsection implements with two explicit adapter branches.

\subsection{Dual-Adapter Architecture}
\label{subsec:architecture}

Implementing the decomposition requires an architecture that produces both $\mathbf{v}_0$ and $\mathbf{g}$ in a single forward pass with minimal parameter overhead.

We freeze the pretrained diffusion backbone and attach two Low-Rank Adaptation (LoRA) adapters~\citep{hu2022lora}, one for each component.
Each adapter computes a low-rank residual from the backbone's hidden representations,
\begin{equation}
  \mathbf{r} = \mathbf{B}\mathbf{A}\,\mathbf{h}_{\text{base}},
  \label{eq:lora}
\end{equation}
where $\mathbf{A} \in \mathbb{R}^{r \times d}$ and $\mathbf{B} \in \mathbb{R}^{d \times r}$ with rank $r \ll d$.
The two branches do not keep separate hidden states. The frozen backbone computes one hidden-state stream per step, both adapters read it at the attention projections, and each branch accumulates its low-rank residuals into its own velocity-shaped output instead of writing them back, so backbone computation runs once per step and $\text{NFE} = N$ is exact.
The guided velocity is then assembled as $\mathbf{v} = \mathbf{v}_0 + s \cdot \mathbf{g}$ per Equation~\ref{eq:decomposition}.
Because $s$ multiplies $\mathbf{g}$ \emph{after} the forward pass, it remains adjustable at inference without retraining.
The full interface is thus preserved with only $2.5\%$ additional parameters per adapter ($5\%$ total, rank $r{=}64$).

The architecture makes the two fields available, but does not determine how training pressure is distributed across them.
A component-wise loss can underweight guidance errors that inference amplifies by $s$, while composed-output training at a single scale leaves the split underdetermined.
Section~\ref{subsec:training} addresses both risks with a unified multi-scale distillation objective.

\subsection{Unified Distillation}
\label{subsec:training}

\textbf{Training-Pressure Mismatch.}
The dual-adapter architecture exposes both $\mathbf{v}_0$ and $\mathbf{g}$, but a naive component-wise objective would optimize the wrong error.
If the two outputs are supervised independently, base-flow and guidance residuals receive equal loss weight even though inference uses the composed velocity $\hat{\mathbf{v}}_0 + s \cdot \hat{\mathbf{g}}$.
A residual error in $\hat{\mathbf{g}}$ is therefore magnified by $s$ at sampling time, whereas a residual error in $\hat{\mathbf{v}}_0$ enters with unit weight.
Because the backbone is frozen and each adapter has fixed rank, this mismatch determines which residuals receive training pressure within a fixed budget.
At large $s$, equal component weighting under-prioritizes the guidance errors most likely to produce artifacts.
Appendix~\ref{app:gradient} formalizes this capacity-allocation analysis.

\textbf{Unified Distillation Objective.}
To align training with the deployed interface, \method supervises the composed output rather than the two components independently.
For each training tuple $(\mathbf{x}_t, c, t, s)$, a frozen CFG teacher provides the two-pass target $\mathbf{v}_{\text{cfg}}^{\text{teacher}}$, and the student minimizes
\begin{equation}
  \mathcal{L} = \mathbb{E}\left[\left\|\hat{\mathbf{v}}_0 + s \cdot \hat{\mathbf{g}} - \mathbf{v}_{\text{cfg}}^{\text{teacher}}\right\|^2\right].
  \label{eq:distill_loss}
\end{equation}
This objective gives both adapters the same composed residual while the guidance branch inherits the chain-rule factor $s$.
As a result, high-scale examples place stronger pressure on $\hat{\mathbf{g}}$: Appendix~\ref{app:gradient} shows that the guidance adapter's squared gradient norm carries a factor $s^2$ relative to the base adapter, and verifies empirically the Jacobian-comparability condition behind this factor.
The loss therefore allocates capacity according to inference-time error importance without a manually tuned component balance.

\textbf{Identifiability and Curriculum.}
Composed-output supervision also creates a uniqueness question: at a single guidance scale, many $(\mathbf{v}_0, \mathbf{g})$ pairs can produce the same sum.
Scale variation resolves this degeneracy.
Under realizability, Appendix~\ref{app:gradient} proves that training on at least two distinct values of $s$ for the same input $(\mathbf{x}_t, c, t)$ uniquely recovers the teacher decomposition (Proposition~\ref{prop:identifiability}), a condition met by sampling $s$ independently of the input.
We use this observation as the organizing principle for a three-phase curriculum: Phase~1 first stabilizes the composed prediction at fixed $s$, Phase~2 expands to $s \sim \mathcal{U}[1, s_{\max}]$ to identify the split, and Phase~3 samples the $(N, s)$ solver regimes used at inference.
This schedule introduces scale and discretization difficulty only after the composed prediction is stable, which avoids the convergence instability seen when Phase~1 is skipped and the single-scale overfitting seen when Phase~2 is skipped (Appendix~\ref{app:curriculum_ablation}).

\textbf{Remaining Gap.}
Unified distillation aligns training pressure with the composed inference-time error, but it does not remove the amplification caused by high guidance scales.
At extreme values of $s$, residual errors in $\hat{\mathbf{g}}$ are multiplied during inference and can produce visible artifacts.
Error amplification of this kind is an inference-time phenomenon, and the next subsection provides an inference-time solution.

\vspace{-1.0em}
\subsection{Scale-Adaptive Guidance Modulation}
\label{subsec:guidance_modulation}
\vspace{-0.5em}

As previewed at the close of Section~\ref{subsec:training}, the guided velocity $\mathbf{v}_0 + s \cdot \mathbf{g}$ multiplies prediction errors in $\mathbf{g}$ by the scale factor $s$ at inference time.
In the high-scale, low-step regime ($s > 4$, $N \leq 4$), even small guidance errors can produce visible artifacts after amplification by $s$.
We address this inference-time error with three lightweight multipliers that target where errors matter, how strongly they are scaled, and whether guidance agrees geometrically with the base flow.
The effective guidance weight becomes $s \cdot w(t) \cdot \alpha(s) \cdot \varphi(a)$, yielding the modulated velocity
\begin{equation}
  \mathbf{v}_{\text{out}}(t) = \mathbf{v}_0 + s \cdot w(t) \cdot \alpha(s) \cdot \varphi(a) \cdot \mathbf{g}.
  \label{eq:modulated_velocity}
\end{equation}
Writing $\gamma \triangleq w(t)\,\alpha(s)\,\varphi(a) \in (0, 1]$, modulation shrinks the amplified guidance error from $s\,\boldsymbol{\epsilon}_g$ to $s\gamma\,\boldsymbol{\epsilon}_g$ but also moves the output away from the CFG velocity by $s(1-\gamma)\,\mathbf{g}$, a trade-off that Appendix~\ref{app:error} states exactly.

\textbf{Temporal Modulation $w(t)$.}
Prior work~\citep{kynkaanniemi2024interval} observed that guidance is most useful in a mid-timestep interval, whereas endpoint guidance can harm quality or add little benefit.
We encode this pattern with a smooth window over $t$ whose width narrows as $s$ grows.
Low scales therefore retain broad guidance, while high scales suppress endpoint guidance where amplification is most severe.
Appendix~\ref{app:hyperparams} gives the exact Gaussian form and hyperparameters.
This scale-adaptive schedule extends fixed-interval guidance~\citep{kynkaanniemi2024interval} and beta-distribution schedules~\citep{malarz2025betacfg}.

\textbf{Scale Dampening $\alpha(s)$.}
Even after temporal windowing narrows where guidance is active, high guidance scales still enlarge residual guidance-direction errors.
To counteract this global scale growth, a monotonically decreasing function $\alpha(s)$ reduces the effective guidance magnitude as $s$ increases.
The function engages only above a threshold $s_{\text{thresh}}$, so dampening leaves low-$s$ generation unchanged.
Appendix~\ref{app:hyperparams} specifies the functional form.

\textbf{Alignment Gating $\varphi(a)$.}
The explicit decomposition into $\mathbf{v}_0$ and $\mathbf{g}$ reveals per-step geometric relationships unavailable to single-output methods.
The alignment $a = \cos(\mathbf{v}_0, \mathbf{g})$ from Section~\ref{subsec:decomposition} measures whether conditioning reinforces or opposes the base flow.
We attenuate guidance only when $a < 0$, where guidance opposes the base flow and is more prone to artifacts, and leave well-aligned steps at full strength.
Because \method makes $a$ available at every step, this correction requires no additional forward passes.
APG~\citep{sadat2024apg} performs a related projection, but it requires two passes and decomposes relative to $\mathbf{v}_{\text{cond}}$ rather than the base flow. Appendix~\ref{app:hyperparams} gives our exact gate.
Only $\varphi(a)$ requires the split, since any method with a scale knob can mimic $w(t)$ and $\alpha(s)$ (Section~\ref{subsec:ablations}).

\textbf{Two-Regime Design.}
This bias is intentional.
At or below $s_{\text{thresh}}$, $\alpha(s)$ and the warp of Section~\ref{subsec:timewarp} are inactive and the window is broad, so \method tracks CFG closely (within 0.10 FID at $(25, 2)$).
Above it, CFG's own dynamics degrade and \method departs from them deliberately: applying windowing and dampening to exact two-pass CFG, which has no prediction error, lowers its FID from 53.84 to 46.02 at $(4, 8)$ (Appendix~\ref{app:mechanism_transfer}).

Velocity-field modulation alone, however, does not address every failure case.
At extreme $(N, s)$ corners, the solver's uniform time grid undersamples the part of the trajectory where the guided velocity changes most rapidly.
Section~\ref{subsec:timewarp} addresses this discretization-side failure by redistributing the fixed evaluations.

\vspace{-1.0em}
\subsection{Reducing Dead Zones via Time-Grid Warping}
\label{subsec:timewarp}
\vspace{-0.7em}

\textbf{Dead Zones.}
Certain $(N, s)$ configurations produce sharply degraded generation quality even after velocity-field modulation.
The baseline DPM-Solver schedule distributes timesteps uniformly across $[0, 1]$, whereas the guided trajectory changes most rapidly in a narrower mid-timestep region.
When the fixed grid spends too many evaluations near smoother endpoints and too few in this high-curvature region, the resulting failure is better explained as temporal under-resolution than as another velocity-field error.
We call these configurations \emph{dead zones}.

\textbf{Scale-Dependent Time Warping.}
To address dead zones without adding evaluations or retraining, \method redistributes the existing $N$ solver calls.
A monotone schedule map $\tau$ transforms the uniform solver indices $\{i/N\}_{i=0}^{N}$ into non-uniform timesteps $\{t_i=\tau(i/N)\}_{i=0}^{N}$.
Its dilation weight $\omega(t)$ is chosen to be small in the mid-timestep region and larger near trajectory endpoints, so the solver takes finer steps where guidance-induced curvature is highest and coarser steps where the trajectory is smoother.
The warp magnitude increases only above the same high-scale threshold used by the dampening function in Section~\ref{subsec:guidance_modulation}, and at lower guidance scales the schedule remains uniform.
Appendices~\ref{app:timewarp} and~\ref{app:hyperparams} give the exact construction, monotonicity proof, and hyperparameter values.

\textbf{Properties.}
Because the warping operates at the ODE solver level, orthogonal to the velocity-field modulations in Section~\ref{subsec:guidance_modulation}, it requires no additional forward passes and can be precomputed once per $(N, s)$ configuration with negligible cost.
Because $\tau$ only moves the solver's timesteps, warping adds no bias to the velocity field (Appendix~\ref{app:timewarp}) and applies to any sampler.
Together, modulation and warping stabilize \emph{what} velocity is computed and \emph{where} the solver evaluates it, while preserving the one-pass inference interface.

\vspace{-0.6cm}
\section{Experiments}
\label{sec:experiments}
\vspace{-1.0em}
\subsection{Experimental Setup}
\label{subsec:exp_setup}
\vspace{-0.5em}

\textbf{Backbone, Dataset, and Protocol.}
We evaluate \method on class-conditional ImageNet 256$\times$256 using DiT-XL/2~\citep{peebles2023dit} as the backbone.
The frozen CFG teacher runs $N_{\text{teacher}} = 250$ steps with DPM-Solver~\citep{lu2022dpmsolver}.
We generate 50K images per configuration following the ADM evaluation suite~\citep{dhariwal2021beatgans} across $N \in \{2, 4, 8, 16, 25\}$ and $s \in \{1.0, 2.0, 4.0, 6.0, 8.0\}$, yielding 25 configurations.
Each configuration is evaluated over three independent 50K-image seed lists, and table entries report mean FID-50K.
We also report Precision, Recall~\citep{kynkaanniemi2019precision}, class adherence (top-1 accuracy of a frozen ResNet-50~\citep{he2016resnet} on the conditioning class), and CFG-Match Accuracy, defined as the percentage of configurations satisfying $\mathrm{FID}_{\method} \leq \mathrm{FID}_{\mathrm{CFG}} + \delta$, where $\delta = \max(2.0,\; 0.10 \times \mathrm{FID}_{\mathrm{CFG}})$. Configurations where \method improves over CFG are counted as matches.
The absolute floor of $2.0$ handles low-FID regimes where sub-unit differences are not visually meaningful, while the $10\%$ relative term scales the tolerance in harder regimes.
The criterion is one-sided by design, since modulation deliberately departs from CFG at high $s$ (Section~\ref{subsec:guidance_modulation}).

\begin{wraptable}[16]{r}{0.50\textwidth}
  \vspace{-1.3em}
  \caption{FID-50K ($\downarrow$) on ImageNet 256$\times$256 with DiT-XL/2. Shaded rows use two passes, unshaded rows one. AGD-MultiScale appears in Table~\ref{tab:contribution_isolation} and Appendix~\ref{app:agd_multiscale}, and full statistics and grid in Appendices~\ref{app:fid_stds}, \ref{app:precision_recall}, and~\ref{app:full_grid}.}
  \label{tab:main_results}
  \centering
  \scriptsize
  \setlength{\tabcolsep}{2pt}
  \resizebox{\linewidth}{!}{%
  \begin{tabular}{@{}lccccccc@{}}
    \toprule
    Method & NFE & \shortstack{ms/img\\[-2pt]\scriptsize($N{=}25$)} & \multicolumn{2}{c}{$s{=}2$ (easy)} & $s{=}4$ & \multicolumn{2}{c}{$s{=}8$ (hard)} \\
    \cmidrule(lr){4-5} \cmidrule(lr){6-6} \cmidrule(lr){7-8}
     & & & $N{=}25$ & $N{=}4$ & $N{=}8$ & $N{=}25$ & $N{=}4$ \\
    \midrule
    \rowcolor{blue!6}[0pt][0pt] CFG & $2N$ & 843 & 2.97 & 9.24 & 11.72 & 21.47 & 53.84 \\
    \rowcolor{blue!6}[0pt][0pt] CFG-Zero* & $2N$ & 854 & 2.81 & 8.93 & 11.14 & 20.67 & 52.67 \\
    \rowcolor{blue!6}[0pt][0pt] CFG + Interval & $2N$ & $\sim$843 & 2.84 & 8.87 & 11.21 & 18.23 & 46.51 \\
    \rowcolor{blue!6}[0pt][0pt] APG & $2N$ & 852 & 3.21 & 10.12 & 12.84 & \underline{13.56} & \underline{39.21} \\
    \rowcolor{blue!6}[0pt][0pt] Autoguidance & $2N$ & 891 & \underline{2.71} & \underline{8.61} & \underline{10.83} & 18.72 & 48.35 \\
    \midrule
    AGD (fixed $s{=}2$) & $N$ & 432 & \textbf{3.04} & 10.08 & 25.81 & 44.92 & 82.17 \\
    \textbf{\method} & $N$ & 437 & 3.07 & \textbf{9.47} & \textbf{10.98} & \textbf{19.28} & \textbf{48.95} \\
    \bottomrule
  \end{tabular}
  }
  \vspace{-0.5em}
\end{wraptable}
\vspace{-0.5em}

\textbf{Baselines.}
We compare against seven methods spanning two-pass refinements and single-pass compression.
(1)~\textbf{CFG}~\citep{ho2022cfg} (two-pass oracle),
(2)~\textbf{CFG + Interval}~\citep{kynkaanniemi2024interval} (two-pass, mid-timestep guidance),
(3)~\textbf{CFG-Zero*}~\citep{fan2025cfgzero} (training-free, two-pass),
(4)~\textbf{APG}~\citep{sadat2024apg} (orthogonal projection, two-pass),
(5)~\textbf{Autoguidance}~\citep{karras2024autoguidance} (two-pass, smaller unconditional model),
(6)~\textbf{AGD-MultiScale}, the primary single-pass baseline, an opaque adapter following AGD~\citep{sadat2025agd} that receives $s$ as an input, trained with $s \sim \mathcal{U}[1, 8]$ at matched capacity (rank 128, 34M parameters) and with the same teacher, curriculum, and optimizer as \method, and
(7)~\textbf{AGD (fixed $s{=}2$)}, a smaller (14M) variant trained at one scale.
Two-pass baselines run official implementations with default hyperparameters, and we train both AGD variants since no DiT-XL/2 AGD checkpoint is public (Appendix~\ref{app:baseline_protocol}).
Appendix~\ref{app:gft} also compares against the released GFT checkpoint.

\textbf{Training Details.}
Both LoRA adapters use rank $r = 64$ ($5\%$ total overhead), and training follows the three-phase curriculum from Section~\ref{subsec:training}.
The full run takes 48 GPU-hours on 8$\times$A100.
Appendices~\ref{app:implementation_details} and~\ref{app:hyperparams} give optimizer settings, phase lengths, LoRA placement, teacher construction, solver settings, and validation protocol.
Modulation and warping hyperparameters are tuned on 5K-image validation samples, while all reported FID values use independent 50K-image generations.
These are the five configurations of Table~\ref{tab:main_results}, and the other 20 grid configurations were never used for selection.

\vspace{-0.5em}
\subsection{Main Results}
\label{subsec:main_results}
\vspace{-0.5em}

\method uses a single forward pass ($\text{NFE} = N$), while CFG and the two-pass baselines require $\text{NFE} = 2N$.
Table~\ref{tab:main_results} reports FID-50K at five representative $(N, s)$ configurations spanning easy to extreme regimes.

\begin{wrapfigure}[15]{r}{0.38\textwidth}
  \centering
  \vspace{-1.0em}
  \includegraphics[width=\linewidth]{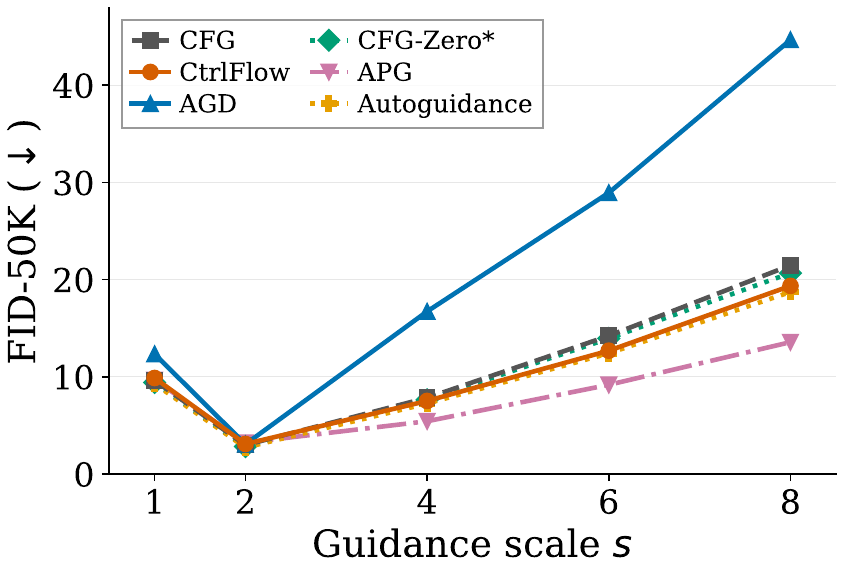}
  \vspace{-1.5em}
  \caption{FID-50K vs.\ scale at $N{=}25$. \method tracks CFG while the fixed-scale AGD variant (``AGD'') drifts past its training scale $s=2$.}
  \label{fig:fid_vs_scale}
  \vspace{-1.0em}
\end{wrapfigure}

At $s{=}2$, where the fixed-scale AGD variant is trained, the tested single-pass methods stay close to CFG, so the easy regime is not the main separator. \method falls within 0.10 FID at $N{=}25$ and within 0.23 at $N{=}4$.
The stronger tests are scale generalization and high-scale control.
The fixed-scale variant degrades once $s$ leaves its training scale (25.81 at $(8, 4)$, 82.17 at $(4, 8)$), and its Recall of 0.14 at $(25, 8)$, against 0.32 for CFG and 0.35 for \method, indicates mode dropping (Appendix~\ref{app:precision_recall}).
Scale-conditioned training removes most of this failure, yet AGD-MultiScale still trails \method at every tested configuration, by up to 4.14 FID at $(25, 8)$, and matches CFG in 20 of 25 grid configurations against 24 of 25 (Appendix~\ref{app:agd_multiscale}).
\method also improves over CFG at $(8, 4)$ (10.98 versus 11.72) and reaches 19.28 and 48.95 at $s{=}8$ with half the forward evaluations.
Class adherence survives the damping: \method's top-1 accuracy stays within 2 points of CFG's at every tested configuration (75.9\% versus 77.8\% at $(4, 8)$), while AGD-MultiScale falls to 57.3\% (Appendix~\ref{app:class_adherence}).
Against the released GFT checkpoint at matched NFE, \method is 0.27 FID worse at $s{\approx}1$ and 0.11 to 2.12 FID better for $s{\approx}2$ to 8, with a frozen backbone instead of full fine-tuning (Appendix~\ref{app:gft}).

Among two-pass methods, APG gives the best high-$s$ FID through orthogonal projection, but it retains $2N$ cost and does not provide a single-pass base-flow/guidance interface.
Among CFG-style methods, \method reaches within 1.05 FID of CFG + Interval at $(25, 8)$ with half the forward evaluations while retaining the decomposition interface.

Figure~\ref{fig:fid_vs_scale} extends this comparison across the full guidance-scale range at $N{=}25$.
The fixed-scale AGD curve climbs sharply beyond its training scale of $s = 2$, whereas \method tracks CFG with half the forward evaluations.
Appendix Figure~\ref{fig:fid_vs_nfe} plots FID against NFE at $s{=}4$ and shows the same compute-quality pattern.

Across the full 25-configuration grid (Appendix~\ref{app:full_grid}), \method achieves $96\%$ CFG-match accuracy (24 of 25 within $\delta$) and a $1.93\times$ wall-clock speedup at $N{=}25$.
This match count is stable across three generation seeds (Appendix~\ref{app:fid_stds}), and all 20 configurations never used for tuning match except $(2, 8)$.
Beyond the training range of $s \le 8$, \method remains competitive, outperforming CFG at $s{=}10$ and trailing by 3.1 FID at $s{=}12$ (Appendix~\ref{app:scale_extrapolation}), and qualitative samples appear in Appendix~\ref{app:qualitative}.

\vspace{-1.0em}
\subsection{Ablation Studies}
\label{subsec:ablations}
\vspace{-0.5em}

\begin{wraptable}[12]{r}{0.50\textwidth}
  \centering
  \vspace{-1.8em}
  \caption{Contribution isolation, FID-50K ($\downarrow$). Core uses dual adapters and unified distillation, and +Mod/+Warp add stabilization. AGD-MultiScale is a matched-budget, scale-conditioned opaque distillation baseline.}
  \label{tab:contribution_isolation}
  \scriptsize
  \setlength{\tabcolsep}{2.5pt}
  \begin{tabular}{@{}lccc@{}}
    \toprule
    Variant & $(25, 2)$ & $(8, 4)$ & $(4, 8)$ \\
    \midrule
    \rowcolor{gray!10}[0pt][0pt] AGD (fixed $s{=}2$) & 3.04 & 25.81 & 82.17 \\
    \rowcolor{gray!10}[0pt][0pt] AGD-MultiScale ($r{=}128$) & 3.09 & 11.94 & 52.87 \\
    \midrule
    \method-Core & 3.14 & 11.48 & 73.41 \\
    \method{}+Mod & 3.12 & 11.21 & 63.42 \\
    \method{}+Warp & 3.15 & 11.35 & 59.83 \\
    \rowcolor{blue!6}[0pt][0pt] \textbf{\method-Full} & \textbf{3.07} & \textbf{10.98} & \textbf{48.95} \\
    \bottomrule
  \end{tabular}
\end{wraptable}

\textbf{Contribution Isolation.}
Table~\ref{tab:contribution_isolation} separates the value of the explicit decomposition from the benefits of multi-scale training by building up from the core architecture.
\method-Core uses only the dual-adapter decomposition with unified distillation, omitting modulation and warping.
As a controlled comparison, AGD-MultiScale trains a single opaque adapter that receives $s$ as an input, with the same multi-scale curriculum and matched parameter budget (rank 128, 34M parameters).
At moderate scales the two are close: AGD-MultiScale reaches FID 11.94 versus \method-Core's 11.48 at $(8,4)$.
At the extreme $(4,8)$, however, AGD-MultiScale achieves 52.87 while \method-Core reaches only 73.41, so the decomposition constraint alone does not provide a raw optimization advantage.
The picture reverses once the interface is exploited, with \method-Full reaching 48.95.

\textbf{Interface versus Transferable Mechanisms.}
Because windowing, dampening, and warping are not specific to the decomposition, we also give them to the baselines, feeding the opaque student a time-varying scale input $s_{\text{eff}}(t) = s\,w(t)\,\alpha(s)$ (Appendix~\ref{app:mechanism_transfer}).
They improve AGD-MultiScale by 0.7 to 2.3 FID, yet \method-Full stays 1.3 to 2.1 FID lower at all four tested configurations, including two never used for tuning, and removing only the alignment gate, the one mechanism that needs $\mathbf{v}_0$ and $\mathbf{g}$, costs 0.6 to 2.4 FID.
The decomposition's value is therefore the interface it opens, not a raw quality gap.

\textbf{Further Ablations.}
Four additional studies (Appendices~\ref{app:ablation_component} to~\ref{app:rank_ablation}) confirm that the mechanisms contribute independently and that the three-phase curriculum is important for stable high-scale performance.
Unified distillation produces the largest single effect (FID degrades by 19.5 at $(4, 8)$ when removed).
The three modulation sub-components are complementary, with each pair outperforming both of its components, and rank 64 sits at the knee of the cost-quality curve.
A conditional-field anchor, $\mathbf{v}_{\text{cond}} + (s-1)\,\mathbf{g}$, matches Core within noise but trails Full by 1.1 to 2.6 FID (Appendix~\ref{app:anchor_ablation}).
Dead-zone analysis and modulation/warping visualizations appear in Appendices~\ref{app:deadzone}, \ref{app:modulation_profiles}, and~\ref{app:time_warp_vis}.

\vspace{-0.5em}
\subsection{Decomposition Fidelity and Interpretability}
\label{subsec:interpretability}
\vspace{-0.5em}

\begin{wraptable}[9]{r}{0.42\textwidth}
  \centering
  \vspace{-1.5em}
  \caption{Component swap, FID-50K ($\downarrow$).}
  \label{tab:component_swap}
  \footnotesize
  \setlength{\tabcolsep}{3.5pt}
  \begin{tabular}{@{}lcc@{}}
    \toprule
    Variant & $(25,2)$ & $(25,8)$ \\
    \midrule
    CFG teacher & 2.97 & 21.47 \\
    Swap $\hat{\mathbf{g}}$ & 3.07 & 22.31 \\
    Swap $\hat{\mathbf{v}}_0$ & 3.01 & 21.74 \\
    \method & 3.07 & 19.28 \\
    \bottomrule
  \end{tabular}
  \vspace{-2.5em}
\end{wraptable}
The explicit decomposition into $\mathbf{v}_0$ and $\mathbf{g}$ enables direct inspection of what guidance changes, a capability unavailable to opaque single-pass methods that predict only the composed guided velocity.
Before using this channel for visualization, we verify that the learned components are faithful to the teacher decomposition in two ways.
Cosine similarity to the teacher-generated ground truth exceeds $0.83$ across all timesteps and peaks at 0.97 for $\hat{\mathbf{v}}_0$ at $t{=}0.5$ (Appendix~\ref{app:decomp_fidelity}).
We then test functional equivalence: each learned component replaces its teacher counterpart in a standard two-pass CFG pipeline, and we measure whether generation quality is preserved.
Table~\ref{tab:component_swap} reports the results.

At $(25,2)$, swapping in $\hat{\mathbf{g}}$ raises FID by only 0.10, while swapping in $\hat{\mathbf{v}}_0$ raises it by 0.04.
At $(25,8)$, the gaps grow to 0.84 for $\hat{\mathbf{g}}$ and 0.27 for $\hat{\mathbf{v}}_0$ because guidance errors are amplified by $s$.
The full \method model reaches 19.28 at $(25,8)$, outperforming both swap variants and CFG itself.
This gain reflects the deliberate high-scale departure from CFG (Appendix~\ref{app:bias_error}), not a closer CFG approximation.
Having verified that the split is both geometrically and functionally meaningful, we next visualize what the two components reveal.

\begin{wrapfigure}{r}{0.48\textwidth}
  \centering
  \vspace{-1.3em}
  \includegraphics[width=\linewidth]{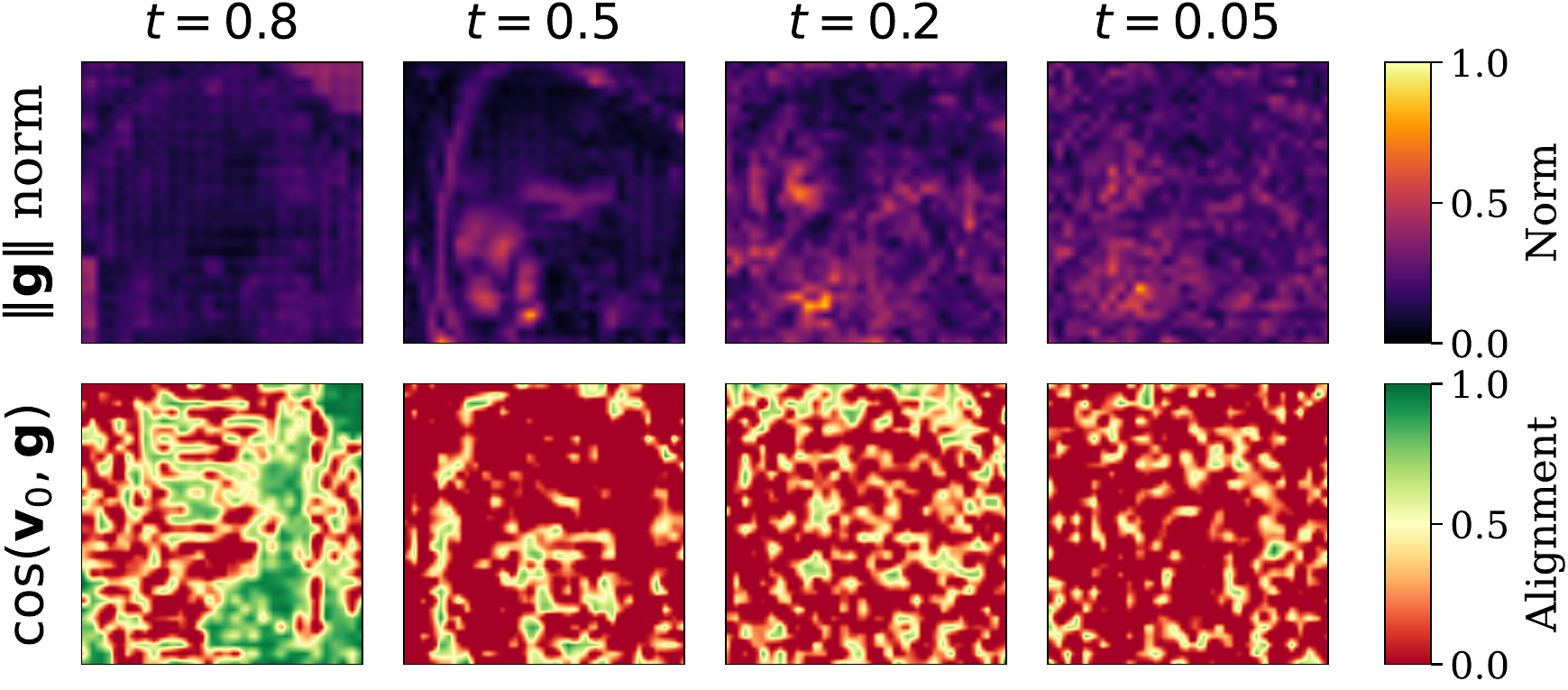}
\vspace{-1.5em}
  \caption{Decomposition maps for \textit{golden retriever}. Guidance norm becomes spatially focused as structure emerges, while $\cos(\mathbf{v}_0,\mathbf{g})$ shifts from near-orthogonal to aligned. Extended classes appear in Appendix~\ref{app:interpretability_extended}.}
  \vspace{-0.5em}
  \label{fig:interpretability}

\end{wrapfigure}

Figure~\ref{fig:interpretability} uses this verified decomposition to visualize where and when guidance acts on a representative ImageNet class (\textit{golden retriever}).
The top row shows that guidance is spatially diffuse at high noise but concentrates by later timesteps on class-discriminative regions such as object boundaries, fur texture, and facial features, while background regions receive little signal.
The bottom row shows the complementary geometric pattern: alignment between $\mathbf{v}_0$ and $\mathbf{g}$ starts near zero and increases toward $1$ as the trajectory converges.
This progression is not imposed explicitly. It emerges from unified distillation and supplies the signal used by the alignment gate $\varphi(a)$ in Section~\ref{subsec:guidance_modulation}.
The same spatial and temporal patterns appear for butterfly and volcano classes in Appendix~\ref{app:interpretability_extended}, suggesting that the learned split is not specific to the representative class shown here.
Quantitatively, guidance-norm maps on noised ImageNet validation images point to the annotated object with accuracy 0.81, above Grad-CAM~\citep{selvaraju2017gradcam} (0.74) and on par with the two-pass teacher's maps (0.80) (Appendix~\ref{app:localization}).

\vspace{-0.7em}
\subsection{Cost and Generalization}
\label{subsec:cost}
\label{subsec:cross_backbone}
\vspace{-0.5em}

The interpretability and stabilization benefits of the decomposition come at modest cost.
\method adds 34M parameters ($5\%$ of DiT-XL/2) and requires 48 GPU-hours of training, but at $N{=}25$ it reduces latency from 843\,ms per image for CFG to 437\,ms, and Appendix Table~\ref{tab:cost} compares all methods.
The speedup is $1.93\times$ rather than $2\times$ because the adapters and modulation add 15.5\,ms to the 421.5\,ms backbone pass (Appendix~\ref{app:cost}).
To test transfer beyond DiT-XL/2, we also train \method on DiT-L/4 with the same protocol.
On DiT-L/4, \method stays within 0.19 FID of CFG at $(25, 2)$ and outperforms CFG at the tested high-scale settings, with full cross-backbone and decomposition-fidelity tables in Appendix~\ref{app:cross_backbone}.
A Stable Diffusion 2.1~\citep{rombach2022ldm} text-to-image pilot keeps the ImageNet ordering: at $N{=}25$ and $s{=}7.5$, \method reaches COCO FID-10K 10.41 and CLIPScore 0.304, between two-pass CFG (9.82, 0.309) and a scale-conditioned opaque student (11.37, 0.300) (Appendix~\ref{app:sd21}).

\vspace{-0.5em}
\section{Conclusion}
\label{sec:conclusion}
\vspace{-1.0em}

\method distills CFG into a single forward pass that preserves the base-flow and guidance-direction interface with two LoRA adapters, keeping $s$ adjustable without retraining.
With unified distillation, scale-adaptive modulation, and time-grid warping, it matches CFG in 24 of 25 $(N, s)$ configurations with a $1.93\times$ speedup at $N{=}25$ on DiT-XL/2, ahead of a matched scale-conditioned opaque student.
Because the same decomposition supports post-hoc scale control, per-step diagnostics, and error-aware stabilization, the results suggest that guidance efficiency and interpretability need not compete.

\textbf{Limitations.}
Our main evaluation covers two DiT backbones on class-conditional ImageNet $256{\times}256$, text conditioning is covered only by a single-setting Stable Diffusion 2.1 pilot, and video and higher resolutions remain untested.
Modulation hyperparameters were selected on the headline configurations, although the 20 untuned ones show the same pattern (Appendix~\ref{app:hyperparams}), and above $s_{\text{thresh}}$ they make \method match CFG quality rather than reproduce CFG samples.
Decomposition fidelity rests on cosine similarity, component swaps, and localization scores, while Proposition~\ref{prop:identifiability} assumes realizability and global zero loss, so finite-sample recovery remains empirical.

\textbf{Broader Impacts.}
The interface provides per-step inspection that could support content-safety pipelines, but making guided generation cheaper inherits the risks of the underlying technology, including deepfake misuse.
Guidance-aware safety mechanisms monitoring decomposition signals deserve dedicated investigation.

\begin{ack}
This work is supported in part by the U.S. National Science Foundation under Grants CNS-2247892, CNS-2239672, OAC-2319962, CNS-2326597, and CCF-2449989.
We thank Modal for supporting this work through Modal academic research credits, which provided part of the compute used in our experiments.
We thank the anonymous reviewers and the area chair for constructive feedback that improved the baselines, analyses, and positioning of this paper.
The authors declare no competing interests.
\end{ack}

\bibliographystyle{plainnat}
\bibliography{citations}

\appendix

\clearpage
\begin{center}
  {\Large\bfseries Appendices}
\end{center}
\vspace{0.8em}
\startcontents[appendices]
\printcontents[appendices]{l}{1}{\setcounter{tocdepth}{3}}

\clearpage
\section{Extended Related Work}
\label{app:extended_related}

This appendix provides additional detail on the related work summarized in Section~\ref{sec:background}.

\paragraph{Classifier-Free Guidance.}
\citet{dhariwal2021beatgans} introduced classifier guidance, which steers sampling toward higher-fidelity outputs by injecting gradients from a separately trained classifier at the cost of reduced sample diversity.
CFG~\citep{ho2022cfg} removed the need for an external classifier by interpolating between conditional and unconditional score estimates.
CFG++~\citep{chung2025cfgpp} recasts guidance as manifold-constrained optimization and shows that standard CFG displaces samples from the learned data manifold when the guidance scale grows large.

\paragraph{Single-Pass Methods.}
Plug-and-Play Diffusion Distillation~\citep{hsiao2024plug} trains a lightweight external guide at 1\% of the base model's parameter count, keeping the base frozen and cutting CFG cost by roughly half.
Building on this idea, AGD~\citep{sadat2025agd} distills CFG into low-rank adapters on a frozen backbone and gives the adapter the guidance scale as an extra input, sampled over a range during training, so one checkpoint supports post-hoc scale changes. Its output is still a single opaque guided velocity.
Guidance-Free Training~\citep{chen2025gft} reaches a similar one-model interface by fine-tuning the full backbone with a temperature input that sets the guidance strength.
Because the strength enters as an input in both methods, each forward pass returns one guided prediction at one strength, and recovering the guidance direction or a second strength costs further forward passes. GFT's temperature also covers only non-negative scales.
\method instead applies the scale after the forward pass, and Appendix~\ref{app:gft} compares the two directly.
GFT generalizes Condition Contrastive Alignment~\citep{chen2024cca}, which fine-tunes autoregressive visual models to sample without guidance, and guidance-free training has since been adopted downstream, for example in the CFG-free reinforcement-learning pipeline of DiffusionNFT~\citep{zheng2026diffusionnft}.
The guidance-free methods we are aware of keep GFT's input-side interface and return one opaque output per pass, so none of them exposes the base flow and guidance direction separately.
Model Guidance~\citep{tang2025modelguidance} sidesteps CFG altogether by folding the posterior probability of conditions directly into the training loss.
Entropy Rectifying Guidance~\citep{berrada2025erg} substitutes attention-based energy signals for the unconditional prediction.
Independent Condition Guidance~\citep{sadat2025icg} conditions on an independent signal instead of an unconditional one, which removes the need for unconditional training yet still requires two forward passes.

\paragraph{Two-Pass Improvements.}
APG~\citep{sadat2024apg} decomposes the guidance update into parallel and orthogonal components relative to the conditional prediction. It selectively down-weights the parallel component to suppress oversaturation artifacts.
CFG-Zero*~\citep{fan2025cfgzero} refines initialization and rescaling specifically for flow matching models.
Autoguidance~\citep{karras2024autoguidance} swaps the unconditional model for a separately trained, lower-capacity network, incurring additional storage and training cost.
Dynamic Negative Guidance~\citep{koulischer2025dng} ties the guidance scale to the posterior probability at each state. Guidance thereby fades in regions where it offers no measurable benefit.

\paragraph{Distillation and Step Reduction.}
Progressive distillation~\citep{salimans2022progressive} repeatedly halves the number of sampling steps.
\citet{meng2023distillation} extended this approach to CFG-guided models, reaching 1 to 4 step generation. Their scale-conditioned student can adjust $s$ freely at inference time.
Consistency models~\citep{song2023consistency} learn direct mappings from noise to data in a single pass.
Latent consistency models~\citep{luo2023lcm} bring this paradigm into the latent diffusion setting.
\citet{golnari2023loradistill} explored combining LoRA with guidance distillation.
Each of these methods treats the guided velocity field as an opaque regression target.

\paragraph{Guidance and Sampling Schedules.}
\citet{kynkaanniemi2024interval} showed that guidance degrades quality at high noise levels and adds nothing near convergence. Only a middle noise interval benefits from guidance, and restricting it to that window improved ImageNet-512 FID from 1.81 to 1.40.
\citet{malarz2025betacfg} generalized this finding with beta-distribution-shaped schedule curves.
Both approaches rely on fixed schedules that do not vary with $s$.
DPM-Solver~\citep{lu2022dpmsolver} leverages the semi-linear ODE structure of diffusion models for efficient high-order numerical solving.
\citet{sabour2024align} demonstrated that optimizing the timestep schedule itself yields substantial gains in few-step regimes.
\citet{chen2024adaptive} proposed gradient-based optimization of discretization points for similar improvements.
All of these methods assume a fixed guidance configuration and do not adapt their behavior to the guidance scale $s$.
Taken together, prior work either preserves CFG's interface at two-pass cost or reduces the pass count by treating the guided velocity as opaque. \method targets the remaining intersection by preserving the interface in a single pass and adapting its use across scale and time.

\section{Theoretical Analysis}
\label{app:theory}

This appendix collects the formal analyses that justify the optimization and control structure of \method.
Section~\ref{app:gradient} analyzes capacity allocation under the unified loss.
Section~\ref{app:error} formalizes the error amplification mechanism and derives bounds for the modulation mechanisms.
Section~\ref{app:timewarp} proves the formal properties of the time-grid warping.

\paragraph{Scope.}
The analyses in Sections~\ref{app:gradient} through~\ref{app:timewarp} do not prove finite-sample convergence, global optimization of LoRA adapters, or optimality of the specific modulation hyperparameters. The identifiability result (Proposition~\ref{prop:identifiability}) assumes realizability and global zero loss.
The capacity-allocation ratio relies on a Jacobian-comparability condition, which we measure rather than assume (Section~\ref{app:gradient}), and the modulation analysis separates the estimation error that modulation suppresses from the bias that it deliberately introduces (Section~\ref{app:bias_error}).
Empirical ablations in Section~\ref{sec:experiments} validate the actual training and inference setting.

\subsection{Capacity Allocation Analysis}
\label{app:gradient}

This section analyzes the capacity misallocation under the naive dual-loss (Section~\ref{subsec:training}) and shows that the unified distillation objective resolves it.
All gradients are with respect to the trainable adapter parameters only, as the backbone is frozen throughout.

\subsubsection{Setup and Notation}

The pretrained backbone is frozen and produces shared features $\mathbf{h}_{\text{base}}(\mathbf{x}_t, c, t)$.
Let $\theta = (\mathbf{A}_0, \mathbf{B}_0, \mathbf{A}_g, \mathbf{B}_g)$ denote the trainable LoRA adapter parameters.
The base-flow adapter computes $\hat{\mathbf{v}}_0 = f_0(\mathbf{h}_{\text{base}}; \mathbf{A}_0, \mathbf{B}_0)$, while the guidance adapter computes $\hat{\mathbf{g}} = f_g(\mathbf{h}_{\text{base}}; \mathbf{A}_g, \mathbf{B}_g)$.
The teacher provides targets $\mathbf{v}_0^*$ and $\mathbf{g}^*$ from two forward passes of the frozen CFG model.
We write the per-sample errors as $\boldsymbol{\delta}_0 \triangleq \hat{\mathbf{v}}_0 - \mathbf{v}_0^*$ and $\boldsymbol{\delta}_g \triangleq \hat{\mathbf{g}} - \mathbf{g}^*$.

Because the backbone is frozen, $\mathbf{h}_{\text{base}}$ is a fixed input to both adapters for any given sample.
The two adapter parameter sets $(\mathbf{A}_0, \mathbf{B}_0)$ and $(\mathbf{A}_g, \mathbf{B}_g)$ are disjoint.
Gradients with respect to one adapter's parameters do not directly affect the other.
The interaction arises indirectly, since both adapters operate on the same frozen features and must represent their respective targets within the same feature space under a shared rank constraint.

\subsubsection{Objective Mismatch under the Naive Loss}

\begin{claim}[Capacity misallocation under $\mathcal{L}_{\text{naive}}$]
\label{prop:conflict}
Consider the naive dual-loss $\mathcal{L}_{\text{naive}} = \|\boldsymbol{\delta}_0\|^2 + \|\boldsymbol{\delta}_g\|^2$.
Because the backbone is frozen, the gradient decomposes into two independent adapter updates,
\begin{equation}
  \nabla_{\theta_0} \mathcal{L}_{\text{naive}} = 2\,\boldsymbol{\delta}_0^\top \nabla_{\theta_0} \hat{\mathbf{v}}_0, \quad
  \nabla_{\theta_g} \mathcal{L}_{\text{naive}} = 2\,\boldsymbol{\delta}_g^\top \nabla_{\theta_g} \hat{\mathbf{g}},
  \label{eq:naive_grad_decomp}
\end{equation}
where $\theta_0 = (\mathbf{A}_0, \mathbf{B}_0)$ and $\theta_g = (\mathbf{A}_g, \mathbf{B}_g)$.
Each adapter receives equal implicit priority. The loss weights both $\|\boldsymbol{\delta}_0\|^2$ and $\|\boldsymbol{\delta}_g\|^2$ by $1$.
At inference, however, the composed output is $\hat{\mathbf{v}}_0 + s\,\hat{\mathbf{g}}$.
An error $\boldsymbol{\delta}_g$ in the guidance component contributes $s\,\boldsymbol{\delta}_g$ to the composed error, while $\boldsymbol{\delta}_0$ contributes only $\boldsymbol{\delta}_0$.
The naive loss ignores this asymmetry.
Because the adapters have fixed ranks, the loss cannot move parameters between branches. Instead, it determines which residual errors receive stronger gradient pressure within the fixed dual-adapter budget.
Equal weighting provides no mechanism to direct that pressure toward the component whose errors matter $s$ times more at inference.
\end{claim}

\textit{Discussion.}
The preceding claim is a training-pressure argument, not a gradient-conflict argument in the classical multi-task sense~\citep{yu2020pcgrad}.
Because the backbone is frozen and the two adapter parameter sets are disjoint, gradients do not directly interfere.
The mismatch arises because the naive loss weights both adapter errors equally, whereas inference amplifies guidance errors by $s$.
Within the fixed dual-adapter budget, the naive loss provides no signal to direct gradient pressure toward the component that matters more at inference.

\subsubsection{Unified Loss Resolves the Mismatch}

\begin{claim}[Unified gradient as a coherent update]
\label{prop:unified}
Under the unified loss $\mathcal{L} = \|\hat{\mathbf{v}}_0 + s\,\hat{\mathbf{g}} - \mathbf{v}_{\text{cfg}}^{\text{teacher}}\|^2$, the gradients with respect to the adapter parameters are
\begin{equation}
  \nabla_{\theta_0} \mathcal{L} = 2\,\boldsymbol{\delta}_{\text{comp}}^\top \nabla_{\theta_0} \hat{\mathbf{v}}_0, \quad
  \nabla_{\theta_g} \mathcal{L} = 2s\,\boldsymbol{\delta}_{\text{comp}}^\top \nabla_{\theta_g} \hat{\mathbf{g}},
\end{equation}
where $\boldsymbol{\delta}_{\text{comp}} = \hat{\mathbf{v}}_0 + s\,\hat{\mathbf{g}} - \mathbf{v}_{\text{cfg}}^{\text{teacher}}$ is the composed error.
Both adapters are driven by the same error signal $\boldsymbol{\delta}_{\text{comp}}$.
The guidance adapter's gradient carries an additional factor of $s$, so its updates are scaled in proportion to the downstream impact of guidance errors.
There is no decomposition into independently weighted objectives and hence no mismatch between training priority and inference importance.
\end{claim}

\textit{Proof sketch.}
Differentiating $\mathcal{L}$ with respect to $\theta_g$ produces the chain-rule factor $\nabla_{\theta_g}(s\,\hat{\mathbf{g}}) = s\,\nabla_{\theta_g}\hat{\mathbf{g}}$.
The factor $s$ appears because the loss directly penalizes the composed output in which $\hat{\mathbf{g}}$ is scaled by $s$.
Differentiating with respect to $\theta_0$ yields no such factor, since $\hat{\mathbf{v}}_0$ enters the composed output with unit weight.

\subsubsection{Scale-Dependent Capacity Allocation}

\begin{claim}[$s^2$ squared-gradient weighting]
\label{prop:scaling}
Let $J_0 \triangleq \nabla_{\theta_0} \hat{\mathbf{v}}_0$ and $J_g \triangleq \nabla_{\theta_g} \hat{\mathbf{g}}$ denote the adapter Jacobians.
Under the unified loss, the squared gradient norms for each adapter satisfy
\begin{equation}
  \|\nabla_{\theta_0} \mathcal{L}\|^2 = 4\,\|\boldsymbol{\delta}_{\text{comp}}^\top J_0\|^2, \quad
  \|\nabla_{\theta_g} \mathcal{L}\|^2 = 4s^2\,\|\boldsymbol{\delta}_{\text{comp}}^\top J_g\|^2,
\end{equation}
so their ratio is exactly
\begin{equation}
  \frac{\|\nabla_{\theta_g} \mathcal{L}\|^2}{\|\nabla_{\theta_0} \mathcal{L}\|^2} = s^2 \cdot \frac{\|\boldsymbol{\delta}_{\text{comp}}^\top J_g\|^2}{\|\boldsymbol{\delta}_{\text{comp}}^\top J_0\|^2}.
  \label{eq:grad_ratio}
\end{equation}
When the two Jacobians respond comparably to the shared composed error, $\|\boldsymbol{\delta}_{\text{comp}}^\top J_g\| \approx \|\boldsymbol{\delta}_{\text{comp}}^\top J_0\|$, the squared-gradient ratio is $\approx s^2$, which corresponds to a gradient-magnitude ratio of $\approx s$.
At $s = 1$, both adapters receive comparable gradient signal.
At $s = 8$, the guidance adapter's squared gradient norm is ideally $s^2 = 64$ times that of the base adapter ($52.7$ as measured below), a gradient magnitude roughly 7 to 8 times larger, which increases training pressure on the component whose errors are amplified most at inference.
\end{claim}

\textit{Proof sketch and empirical check.}
The $s^2$ factor follows from squaring the chain-rule factor $s$ in $\nabla_{\theta_g}\mathcal{L}$, so Equation~\ref{eq:grad_ratio} holds without assumptions.
The comparability condition is not implied by equal rank alone, because the two adapters map the same composed error through different learned weights, so we measure it rather than assume it.
On matched training states, the measured squared-gradient ratio is 4.3, 15.1, and 52.7 at $s = 2$, 4, and 8, against $s^2 = 4$, 16, and 64.
The Jacobian ratio in Equation~\ref{eq:grad_ratio} therefore stays between 0.82 and 1.08 across the training range, and the $s^2$ approximation holds to within $18\%$.

This $s^2$ weighting is the formal mechanism behind the claim in Section~\ref{subsec:training} that the chain-rule factor $s$ scales the guidance-adapter gradient.
Unlike gradient surgery approaches~\citep{yu2020pcgrad} that project conflicting task gradients post-hoc, the unified loss produces a coherent training signal.
It allocates training pressure proportional to the downstream impact of each component's error.

\textit{Scope of this analysis.}
The $s^2$ weighting shows that the unified loss aligns training pressure with inference-time error importance.
Whether the learned decomposition uniquely recovers the teacher's $\mathbf{v}_0$ and $\mathbf{g}$ is a separate question.
Proposition~\ref{prop:identifiability} below addresses it under a realizability assumption, and Section~\ref{subsec:interpretability} validates it empirically.

\subsubsection{Decomposition Identifiability under Scale Variation}

The preceding analysis shows that the unified loss allocates training pressure proportionally to inference-time importance.
A separate question is whether the loss has a unique minimizer that recovers the true decomposition.
The following proposition shows that it does, provided the training distribution includes variation in $s$.

\begin{proposition}[Identifiability under scale variation]
\label{prop:identifiability}
Let $\mathcal{F}_0$ and $\mathcal{F}_g$ be function classes for the base and guidance adapters.
Suppose:
\begin{enumerate}
  \item[(i)] For almost every input $(\mathbf{x}_t, c, t)$, the training distribution of $s$ given that input has support containing at least two distinct values $s_1 \neq s_2$.
  \item[(ii)] There exist $f_0^* \in \mathcal{F}_0$ and $f_g^* \in \mathcal{F}_g$ such that $\mathbf{v}_{\mathrm{cfg}}(\mathbf{x}_t, c, t, s) = f_0^*(\mathbf{h}) + s \cdot f_g^*(\mathbf{h})$ for all $(\mathbf{x}_t, c, t, s)$.
\end{enumerate}
Then the global minimizer of $\mathcal{L} = \mathbb{E}_{s}\!\left[\|f_0(\mathbf{h}) + s \cdot f_g(\mathbf{h}) - \mathbf{v}_{\mathrm{cfg}}\|^2\right]$ is unique: $f_0 = f_0^*$ and $f_g = f_g^*$.
\end{proposition}

\begin{proof}
At the global minimum, $\mathcal{L} = 0$.
For any sample $(\mathbf{x}_t, c, t)$, define $\boldsymbol{\delta}_0 = f_0(\mathbf{h}) - f_0^*(\mathbf{h})$ and $\boldsymbol{\delta}_g = f_g(\mathbf{h}) - f_g^*(\mathbf{h})$.
The residual $r(s) = \boldsymbol{\delta}_0 + s \cdot \boldsymbol{\delta}_g$ must equal zero for all $s$ in the support.
Evaluating at $s_1$ and $s_2$:
\begin{align}
  \boldsymbol{\delta}_0 + s_1 \cdot \boldsymbol{\delta}_g &= \mathbf{0}, \\
  \boldsymbol{\delta}_0 + s_2 \cdot \boldsymbol{\delta}_g &= \mathbf{0}.
\end{align}
Subtracting yields $(s_1 - s_2) \cdot \boldsymbol{\delta}_g = \mathbf{0}$.
Since $s_1 \neq s_2$, we have $\boldsymbol{\delta}_g = \mathbf{0}$, and substituting back gives $\boldsymbol{\delta}_0 = \mathbf{0}$.
\end{proof}

\textit{Interpretation.}
Condition~(i) must hold for each input, not only over the whole training set, since a schedule that paired each input with a single scale would leave the split undetermined input by input.
Phases~2 and~3 of the curriculum satisfy it because they sample $s$ from a continuous distribution independently of $(\mathbf{x}_t, c, t)$.
Condition~(ii) is the realizability assumption, requiring that the adapter function classes be expressive enough to represent the true components.
In practice, low-rank adapters provide an approximation, and the decomposition is correspondingly approximate.
The proposition shows that the unified loss has the correct inductive bias: when the true components are representable and the training distribution includes scale variation, the global minimizer recovers the teacher decomposition.
In finite-capacity LoRA adapters, this result should be read as an inductive-bias argument rather than a convergence guarantee. The decomposition is correspondingly approximate, and Section~\ref{subsec:interpretability} validates it empirically.

This result also explains why Phase~1 (fixed $s = s_{\text{target}}$) cannot identify the decomposition.
At a single scale, the loss constrains only the sum $f_0(\mathbf{h}) + s_{\text{target}} \cdot f_g(\mathbf{h})$, leaving infinitely many valid $(f_0, f_g)$ pairs.
Scale variation in Phases~2 and~3 breaks this degeneracy.

\subsection{Error Amplification Analysis}
\label{app:error}

This section formalizes the error amplification mechanism described in Section~\ref{subsec:guidance_modulation} and derives bounds that justify the three modulation mechanisms.
It then states the exact decomposition of the modulated output against the CFG velocity, which separates the estimation error that modulation suppresses from the bias that it deliberately introduces (Section~\ref{app:bias_error}).

\subsubsection{Error Model}

Let $\mathbf{v}_0$ and $\mathbf{g}$ denote the true (teacher-generated) base velocity and guidance direction.
The student produces approximations $\hat{\mathbf{v}}_0 = \mathbf{v}_0 + \boldsymbol{\epsilon}_0$ and $\hat{\mathbf{g}} = \mathbf{g} + \boldsymbol{\epsilon}_g$, where $\boldsymbol{\epsilon}_0$ and $\boldsymbol{\epsilon}_g$ are the prediction errors at timestep $t$.
The true guided velocity is $\mathbf{v}^* = \mathbf{v}_0 + s \cdot \mathbf{g}$, and the student's approximation is $\hat{\mathbf{v}} = \hat{\mathbf{v}}_0 + s \cdot \hat{\mathbf{g}}$.

\subsubsection{Per-Step Error Bound}

\begin{proposition}[Per-step velocity error]
\label{prop:perstep}
The velocity error at any timestep $t$ satisfies
\begin{equation}
  \|\hat{\mathbf{v}}(t) - \mathbf{v}^*(t)\| = \|\boldsymbol{\epsilon}_0(t) + s \cdot \boldsymbol{\epsilon}_g(t)\| \leq \|\boldsymbol{\epsilon}_0(t)\| + s\,\|\boldsymbol{\epsilon}_g(t)\|.
  \label{eq:perstep_bound}
\end{equation}
\end{proposition}

\begin{proof}
Direct computation: $\hat{\mathbf{v}} - \mathbf{v}^* = (\mathbf{v}_0 + \boldsymbol{\epsilon}_0 + s(\mathbf{g} + \boldsymbol{\epsilon}_g)) - (\mathbf{v}_0 + s\mathbf{g}) = \boldsymbol{\epsilon}_0 + s\,\boldsymbol{\epsilon}_g$.
The bound follows from the triangle inequality.
\end{proof}

At $s = 1$, both error terms contribute equally.
At $s = 8$, the guidance-error contribution is amplified by a factor of 8 while the base error remains at unit weight.
When guidance error dominates the base-error term, the total trajectory error approaches an $8\times$ increase over the $s = 1$ case.
This asymmetry is the fundamental source of quality degradation at high guidance scales.

\subsubsection{Trajectory Error Accumulation}

\begin{proposition}[Trajectory error over $N$ steps]
\label{prop:trajectory}
Consider an Euler discretization of the ODE $d\mathbf{x}/dt = \mathbf{v}(\mathbf{x}, t)$ with uniform step size $h = 1/N$.
Let $\mathbf{x}_n$ and $\hat{\mathbf{x}}_n$ denote the true and approximate trajectories at step $n$.
If the velocity field is Lipschitz with constant $L$ and the per-step velocity error is bounded by $\|\hat{\mathbf{v}}(t_n) - \mathbf{v}^*(t_n)\| \leq E(s)$ for all $n$, then
\begin{equation}
  \|\hat{\mathbf{x}}_N - \mathbf{x}_N\| \leq \frac{E(s)}{L}\left(e^{L} - 1\right),
  \label{eq:trajectory_bound}
\end{equation}
where $E(s) \leq \bar{\epsilon}_0 + s\,\bar{\epsilon}_g$ with $\bar{\epsilon}_0 = \max_t \|\boldsymbol{\epsilon}_0(t)\|$ and $\bar{\epsilon}_g = \max_t \|\boldsymbol{\epsilon}_g(t)\|$.
\end{proposition}

\begin{proof}
The trajectory bound is a standard result from ODE perturbation theory (Gronwall's inequality).
At step $n+1$:
\begin{equation}
  \hat{\mathbf{x}}_{n+1} - \mathbf{x}_{n+1} = (\hat{\mathbf{x}}_n - \mathbf{x}_n) + h\left(\hat{\mathbf{v}}(\hat{\mathbf{x}}_n, t_n) - \mathbf{v}^*(\mathbf{x}_n, t_n)\right).
\end{equation}
By the Lipschitz condition and the triangle inequality:
\begin{equation}
  \|\hat{\mathbf{x}}_{n+1} - \mathbf{x}_{n+1}\| \leq (1 + hL)\|\hat{\mathbf{x}}_n - \mathbf{x}_n\| + h\,E(s).
\end{equation}
Applying the discrete Gronwall inequality over $N$ steps with $h = 1/N$ yields the bound.
\end{proof}

The trajectory error scales linearly with $E(s) = O(\bar{\epsilon}_0 + s\,\bar{\epsilon}_g)$.
When guidance error dominates ($\bar{\epsilon}_g \gg \bar{\epsilon}_0$), the trajectory error at $s = 8$ approaches $8\times$ the error at $s = 1$.

\subsubsection{How Each Modulation Mechanism Reduces Amplification}

Let $\gamma \triangleq w(t)\,\alpha(s)\,\varphi(a) \in (0, 1]$ denote the combined modulation factor, with $\gamma = 1$ meaning no modulation.
The bounds in this subsection compare the modulated student with the equally modulated teacher field $\mathbf{v}_0 + s\gamma\,\mathbf{g}$.
They therefore measure how much amplified estimation error each mechanism removes. The price of modulation, a shift of that target away from the CFG velocity $\mathbf{v}^*$, is analyzed in Section~\ref{app:bias_error}.

\paragraph{Temporal Modulation $w(t)$.}
Replacing $s$ with $s \cdot w(t)$ in the guided velocity bounds the effective per-step error by $\|\boldsymbol{\epsilon}_0(t)\| + s \cdot w(t) \cdot \|\boldsymbol{\epsilon}_g(t)\|$.
Since $w(t) \approx 0$ at trajectory endpoints (where $\|\boldsymbol{\epsilon}_g\|$ is empirically largest due to high noise or near-convergence), the peak velocity error is reduced.
The trajectory bound becomes
\begin{equation}
  E_w(s) \leq \bar{\epsilon}_0 + s \cdot \max_t\!\left[w(t) \cdot \|\boldsymbol{\epsilon}_g(t)\|\right] \leq \bar{\epsilon}_0 + s \cdot \bar{\epsilon}_g^{\text{mid}},
\end{equation}
where $\bar{\epsilon}_g^{\text{mid}} \leq \bar{\epsilon}_g$ is the maximum guidance error in the mid-timestep window where $w(t)$ is non-negligible.

\paragraph{Scale Dampening $\alpha(s)$.}
Replacing $s$ with $s \cdot \alpha(s)$ caps the effective amplification at $s \cdot \alpha(s) < s$ for $s > s_{\text{thresh}}$.
The dampened scale yields a global reduction, giving $E_\alpha(s) \leq \bar{\epsilon}_0 + s \cdot \alpha(s) \cdot \bar{\epsilon}_g$.

\paragraph{Alignment Gating $\varphi(a)$.}
The alignment $a = \cos(\mathbf{v}_0, \mathbf{g})$ is empirically associated with guidance prediction error in our diagnostics: misaligned steps ($a < 0$) tend to coincide with larger prediction errors, likely because the guidance direction opposes the base flow and creates a harder prediction target.
Applying $\varphi(a) < 1$ at these high-error timesteps reduces the effective error to $s \cdot \varphi(a) \cdot \|\boldsymbol{\epsilon}_g(t)\|$. The modulation ablation (Table~\ref{tab:ablation_modulation}) confirms that the gate improves FID, providing indirect evidence for the correlation.
Unlike $w(t)$ (which operates in the time domain) and $\alpha(s)$ (which operates in the scale domain), $\varphi(a)$ operates in the geometric domain and adapts \emph{per-sample} and \emph{per-timestep} based on the actual velocity-field geometry.

\paragraph{Composed Modulation.}
Under full modulation, the student outputs $\hat{\mathbf{v}}_{\text{out}} = \hat{\mathbf{v}}_0 + s\gamma\,\hat{\mathbf{g}}$, and its estimation error against the modulated teacher field satisfies
\begin{equation}
  \|\hat{\mathbf{v}}_{\text{out}}(t) - (\mathbf{v}_0 + s\gamma\,\mathbf{g})(t)\| \leq \|\boldsymbol{\epsilon}_0(t)\| + s \cdot w(t) \cdot \alpha(s) \cdot \varphi(a(t)) \cdot \|\boldsymbol{\epsilon}_g(t)\|.
  \label{eq:composed_bound}
\end{equation}
Each factor reduces a different component of the error.
Specifically, $w(t)$ suppresses temporal endpoints, $\alpha(s)$ caps the global scale, and $\varphi(a)$ targets the geometric regime where errors are largest.
Because they operate on orthogonal dimensions (time, scale, geometry), their composition is multiplicative, and no single mechanism can substitute for the others.
This analysis is consistent with the ablation results in Table~\ref{tab:ablation_modulation}, which show that every pair of mechanisms outperforms both of its components and the full trio achieves the lowest error.

\subsubsection{Bias-Error Trade-off and Two-Regime Design}
\label{app:bias_error}

Equation~\ref{eq:composed_bound} measures error against the modulated target, but the samples are ultimately judged against what CFG would produce.
The following decomposition makes the gap between the two explicit.

\begin{proposition}[Exact decomposition against the CFG velocity]
\label{prop:bias_error}
With $\hat{\mathbf{v}}_{\text{out}} = \hat{\mathbf{v}}_0 + s\gamma\,\hat{\mathbf{g}}$ and $\mathbf{v}^* = \mathbf{v}_0 + s\,\mathbf{g}$,
\begin{equation}
  \hat{\mathbf{v}}_{\text{out}} - \mathbf{v}^* = \underbrace{\boldsymbol{\epsilon}_0 + s\gamma\,\boldsymbol{\epsilon}_g}_{\text{estimation error}} + \underbrace{s(\gamma - 1)\,\mathbf{g}}_{\text{modulation bias}},
  \qquad
  \|\hat{\mathbf{v}}_{\text{out}} - \mathbf{v}^*\| \leq \|\boldsymbol{\epsilon}_0\| + s\gamma\,\|\boldsymbol{\epsilon}_g\| + s(1-\gamma)\,\|\mathbf{g}\|.
  \label{eq:bias_error}
\end{equation}
\end{proposition}

\begin{proof}
Substituting $\hat{\mathbf{v}}_0 = \mathbf{v}_0 + \boldsymbol{\epsilon}_0$ and $\hat{\mathbf{g}} = \mathbf{g} + \boldsymbol{\epsilon}_g$ gives $\hat{\mathbf{v}}_{\text{out}} - \mathbf{v}^* = \boldsymbol{\epsilon}_0 + s\gamma\,\boldsymbol{\epsilon}_g + s(\gamma - 1)\,\mathbf{g}$.
Since $\gamma \in (0, 1]$, $|\gamma - 1| = 1 - \gamma$, and the bound follows from the triangle inequality.
The identity holds for any value of $\gamma$, including the one computed from the student's own alignment $\hat{a} = \cos(\hat{\mathbf{v}}_0, \hat{\mathbf{g}})$ at inference.
\end{proof}

\paragraph{Interpretation.}
The first two terms are the estimation error, which modulation reduces by shrinking the coefficient on $\boldsymbol{\epsilon}_g$ from $s$ to $s\gamma$.
The third term is a bias, because the output now follows $\mathbf{v}_0 + s\gamma\,\mathbf{g}$ rather than the CFG velocity.
The bias is not small at high scale. At $s = 8$ and the window center, where $w = \varphi = 1$, dampening alone gives $\gamma = \alpha(8) = 0.7$ and a bias of $2.4\,\|\mathbf{g}\|$.
Modulation therefore pays off only if CFG's own high-scale dynamics are worth leaving, which is an empirical question.

\paragraph{Two Regimes.}
The design separates two regimes at $s_{\text{thresh}} = 4$.
At or below the threshold, $\alpha(s) = 1$ and $\lambda(s) = 0$, the temporal window is broad, and $\varphi(a) = 1$ whenever $a \geq 0$, so modulation is mild and \method tracks CFG closely (3.07 versus 2.97 at $(25, 2)$).
Above the threshold, the target departs from CFG deliberately, and three measurements support this choice.
\begin{itemize}
  \item \emph{Zero-error control.} Exact two-pass CFG has $\boldsymbol{\epsilon}_0 = \boldsymbol{\epsilon}_g = \mathbf{0}$, so any change it shows under modulation comes from the bias term alone. Applying $w\alpha$ alone lowers its FID from 53.84 to 46.02 at $(4, 8)$ and from 21.47 to 17.89 at $(25, 8)$, and adding warping, which changes only the discretization, lowers it further to 42.31 and 17.42 (Table~\ref{tab:mechanism_transfer}).
  \item \emph{Component swap.} The unmodulated pipeline gives 21.47 FID at $(25, 8)$ with exact teacher components and 21.74 to 22.31 when one of them is replaced by the learned one, whereas \method-Full reaches 19.28 (Table~\ref{tab:component_swap}), so the gain cannot come from a closer CFG approximation.
  \item \emph{Relative size.} On matched states at $s = 8$, the integrated bias $\int s(1-\gamma)\,\|\mathbf{g}\|\,dt$ exceeds the integrated amplified error $\int s\gamma\,\|\boldsymbol{\epsilon}_g\|\,dt$ by about $3.2\times$, so the dominant effect of modulation at high scale is a deliberate shift of the target.
\end{itemize}
Time-grid warping does not enter Equation~\ref{eq:bias_error}, since it changes where the solver evaluates the field rather than the field itself (Section~\ref{app:timewarp}).
Because the high-scale target is intentionally not CFG, the CFG-Match criterion of Section~\ref{subsec:exp_setup} counts improvements over CFG as matches, which makes it one-sided.

\subsection{Time-Warp Properties}
\label{app:timewarp}

This section proves the formal properties of the time-grid warping mechanism (Section~\ref{subsec:timewarp}) and discusses its relationship to discretization error.

\subsubsection{Monotonicity and Endpoint Preservation}

\begin{proposition}[Well-definedness of $\tau$]
\label{prop:monotone}
Let $\omega\!: [0,1] \to \mathbb{R}_{>0}$ be a continuous, strictly positive function.
Define
\begin{equation}
  \tau(t) = \frac{\int_0^t \omega(u)\,du}{\int_0^1 \omega(u)\,du}.
\end{equation}
Then $\tau$ is a valid time-warping function. It is strictly monotonically increasing, continuously differentiable, and satisfies $\tau(0) = 0$ and $\tau(1) = 1$.
\end{proposition}

\begin{proof}
\emph{Endpoint preservation.}
$\tau(0) = 0/\int_0^1 \omega(u)\,du = 0$ and $\tau(1) = \int_0^1 \omega(u)\,du / \int_0^1 \omega(u)\,du = 1$.

\emph{Monotonicity.}
The derivative is $\tau'(t) = \omega(t) / \int_0^1 \omega(u)\,du$.
Since $\omega(t) > 0$ for all $t \in [0,1]$ and the denominator is a positive constant, $\tau'(t) > 0$ everywhere.
Hence $\tau$ is strictly increasing.

\emph{Smoothness.}
Since $\omega$ is continuous, $\tau$ is continuously differentiable by the fundamental theorem of calculus.
\end{proof}

\begin{corollary}
The \method dilation weight $\omega(t) = 1 + \lambda(s) \cdot d(t)$ satisfies $\omega(t) > 0$ for all $t$ and all $s \geq 0$, since $\lambda(s) \geq 0$ and $d(t) = \mathrm{sigmoid}(\cdot) \in (0, 1)$.
Hence the resulting $\tau$ is always a valid warping function.
\end{corollary}

\begin{remark}[No velocity-field bias]
Because $\tau$ is strictly increasing with $\tau(0) = 0$ and $\tau(1) = 1$, warping only moves the $N$ solver timesteps along the same trajectory interval and leaves the velocity field being integrated unchanged.
It therefore adds no term to the bias-error decomposition of Proposition~\ref{prop:bias_error}, and it affects only the discretization error analyzed below.
\end{remark}

\subsubsection{Discretization Error Reduction}

The following proposition formalizes the intuition that concentrating solver evaluations near high-curvature regions reduces ODE integration error.

\begin{proposition}[Curvature-weighted step allocation]
\label{prop:truncation}
Consider an ODE $d\mathbf{x}/dt = \mathbf{v}(\mathbf{x}, t)$ integrated with Euler's method using step sizes $h_n = t_{n+1} - t_n$.
The local truncation error at step $n$ is
\begin{equation}
  \boldsymbol{e}_n = \frac{h_n^2}{2}\,\frac{d^2\mathbf{x}}{dt^2}\bigg|_{t=t_n} + O(h_n^3) = \frac{h_n^2}{2}\left(\frac{\partial \mathbf{v}}{\partial t} + \frac{\partial \mathbf{v}}{\partial \mathbf{x}}\,\mathbf{v}\right)\bigg|_{t_n} + O(h_n^3).
  \label{eq:local_truncation}
\end{equation}
Let $c_n = \|d^2\mathbf{x}/dt^2|_{t_n}\|$.
For a fixed total number of steps $N$ (so $\sum_n h_n = 1$), reducing $h_n$ in regions with larger $c_n$ decreases the leading-order error.
The exact optimal exponent depends on the aggregate error objective.
Minimizing $\sum_n h_n^2 c_n$ subject to $\sum_n h_n = 1$ gives $h_n \propto c_n^{-1}$, while equidistributing local error magnitudes $h_n^2 c_n$ gives $h_n \propto c_n^{-1/2}$.
Both allocations place smaller steps where curvature is larger.
\end{proposition}

\begin{proof}
The local truncation error follows from Taylor expansion of the exact solution.
The total error is $E = \sum_{n=0}^{N-1} \|\boldsymbol{e}_n\| \approx \frac{1}{2}\sum_n h_n^2 \, c_n$ where $c_n = \|d^2\mathbf{x}/dt^2|_{t_n}\|$.
Minimizing $E = \sum_n h_n^2 c_n$ subject to $\sum_n h_n = 1$ via Lagrange multipliers gives the first-order condition $h_n c_n = \text{const}$, yielding $h_n \propto c_n^{-1}$.
Alternatively, equidistributing local error $h_n^2 c_n = \text{const}$ gives $h_n \propto c_n^{-1/2}$.
Both allocations assign smaller steps to regions with larger trajectory curvature.
\end{proof}

\paragraph{Connection to \method's Warping.}
The second derivative of the guided trajectory $\|d^2\mathbf{x}/dt^2\|$ depends on the velocity field's spatial and temporal curvature.
At high guidance scales, the curvature concentrates in the mid-timestep region where $\|\mathbf{g}\|$ is large and changes rapidly.
At trajectory endpoints, the velocity field is smoother, being nearly unconditional at $t \approx 1$ and nearly converged at $t \approx 0$.

\method's schedule map $\tau$ approximates a curvature-weighted step allocation.
The dilation weight $\omega(t)$ is large at trajectory endpoints, so the schedule map advances quickly there and the solver takes wider steps.
In the mid-timestep region, $\omega(t)$ is small, so the schedule map advances slowly and the solver takes finer steps.
The resulting grid concentrates evaluations where the velocity field curvature is highest.
The warping is not strictly optimal in the sense of Proposition~\ref{prop:truncation}, since it uses a heuristic $d(t)$ rather than the true second derivative.
It nonetheless aligns with the qualitative structure of curvature-weighted allocation and empirically reduces the dead-zone phenomenon (Figure~\ref{fig:deadzone}).
The Euler derivation explains why smaller local steps should be allocated to high-curvature regions.
DPM-Solver has different truncation constants, yet the same qualitative dependence on schedule allocation motivates the heuristic.

\subsubsection{Computational Cost}

The warped schedule $\{t_0, t_1, \ldots, t_N\}$ is computed by evaluating the CDF-like integral $\tau(i/N)$ at $N+1$ points.
The trapezoidal rule on a fine grid of $K$ points yields the following costs.
\begin{itemize}
  \item Evaluate $\omega(t_k)$ at $K$ points, requiring $O(K)$ operations. Each evaluation computes $m(t_k)$, $d(t_k)$, and $\omega(t_k)$, all closed-form expressions.
  \item Compute the cumulative sum $\int_0^{t_k} \omega(u)\,du$ via prefix sum in $O(K)$.
  \item Normalize by $\int_0^1 \omega(u)\,du$ in $O(K)$.
  \item Look up $\tau(i/N)$ by interpolation in $O(N \log K)$ with binary search.
\end{itemize}
With $K = 1000$ (our default), the total precomputation is negligible ($< 1$\,ms) relative to the forward-pass cost ($\approx 17$\,ms per step).
The warped schedule is computed once per $(N, s)$ configuration and cached.

\section{Hyperparameter Specification}
\label{app:hyperparams}

This appendix lists the functional forms and numerical values of every hyperparameter introduced in the methodology.

\subsection{Modulation Functions (Section~\ref{subsec:guidance_modulation})}

\paragraph{Temporal Window $w(t)$.}
The temporal modulation is a Gaussian window,
\begin{equation}
  w(t) = \exp\!\left(-\frac{(t - \mu)^2}{2\,\sigma(s)^2}\right),
  \label{eq:temporal_modulation}
\end{equation}
where $\mu$ centers the window and $\sigma(s)$ controls its width.

\paragraph{Temporal Window Width $\sigma(s)$.}
The width of the Gaussian temporal modulation $w(t)$ (Equation~\ref{eq:temporal_modulation}) decreases linearly with $s$, following
\begin{equation}
  \sigma(s) = \sigma_{\max} - (\sigma_{\max} - \sigma_{\min}) \cdot \frac{\min(s, s_{\max}) - 1}{s_{\max} - 1},
\end{equation}
where $\sigma_{\max} = 0.35$ produces a broad window at $s = 1$ and $\sigma_{\min} = 0.10$ yields a narrow window at $s = s_{\max}$.
The window center remains fixed at $\mu = 0.5$, the midpoint of the $[0,1]$ time interval.

\paragraph{Scale Dampening $\alpha_{\text{damp}}(s)$.}
The dampening function (Section~\ref{subsec:guidance_modulation}) applies a linear decay above the activation threshold, given by
\begin{equation}
  \alpha_{\text{damp}}(s) = 1 - \beta \cdot \frac{s - s_{\text{thresh}}}{s_{\max} - s_{\text{thresh}}},
\end{equation}
where $\beta = 0.3$ controls the maximum dampening.
At $s = s_{\max} = 8$, $\alpha_{\text{damp}} = 0.7$, reducing the effective guidance scale from $8$ to $5.6$.
Exponential and sigmoid alternatives were also tested. Linear decay matched their performance while remaining simpler to implement.

\paragraph{Alignment Gating $\varphi(a)$.}
As defined in Section~\ref{subsec:guidance_modulation}, $\varphi(a) = 1$ for $a \geq 0$ and $\varphi(a) = \max(\varphi_{\min}, 1 + a)$ for $a < 0$, with $\varphi_{\min} = 0.1$.

\subsection{Time-Warp Parameters (Section~\ref{subsec:timewarp})}

\paragraph{Time Score Parameters.}
The time score $m(t) = \exp(-(t - \mu_w)^2 / (2\sigma_w^2))$ uses $\mu_w = 0.5$ and $\sigma_w = 0.25$.
These values are chosen independently of the temporal modulation parameters $\mu$ and $\sigma(s)$. This separation lets the warping respond to a broader region of the trajectory.

\paragraph{Deadness Gate.}
The soft gate $d(t) = \text{sigmoid}(\kappa \cdot (m_0 - m(t)))$ uses threshold $m_0 = 0.3$ and sharpness $\kappa = 10$.
With $m_0 = 0.3$, the gate transitions near $t = 0.15$ and $t = 0.85$, flagging the outer 30\% of the trajectory as dead-zone-prone.

\paragraph{Warp Magnitude.}
$\lambda(s) = \lambda_0 \cdot \max(0, s - s_{\text{thresh}}) / s_{\max}$ with $\lambda_0 = 1.0$.
At $s = 8$, $\lambda(8) = 1.0 \cdot (8 - 4)/8 = 0.5$.
The maximum dilation weight therefore reaches $\omega_{\max} = 1 + \lambda(s) \cdot 1 = 1.5$, because $d(t) \leq 1$.

\subsection{Shared Parameters}

The activation threshold $s_{\text{thresh}} = 4$ is shared by the scale dampening $\alpha(s)$ and the time-warp magnitude $\lambda(s)$.
Below this threshold, both mechanisms remain inactive, leaving low-$s$ generation unmodified.

The training range upper bound $s_{\max} = 8$ normalizes $\sigma(s)$, $\alpha_{\text{damp}}(s)$, and $\lambda(s)$.

\subsection{Summary Table}

\begin{table}[H]
  \caption{Complete hyperparameter specification for \method's modulation and warping mechanisms.}
  \label{tab:hyperparams}
  \centering
  \small
  \begin{tabular}{llcl}
    \toprule
    Symbol & Description & Value & Used in \\
    \midrule
    \multicolumn{4}{l}{\emph{Temporal Modulation}} \\
    $\mu$ & Window center & 0.5 & Eq.~\ref{eq:temporal_modulation} \\
    $\sigma_{\max}$ & Window width at $s{=}1$ & 0.35 & $\sigma(s)$ \\
    $\sigma_{\min}$ & Window width at $s{=}s_{\max}$ & 0.10 & $\sigma(s)$ \\
    \midrule
    \multicolumn{4}{l}{\emph{Scale Dampening}} \\
    $s_{\text{thresh}}$ & Activation threshold & 4.0 & Sec.~\ref{subsec:guidance_modulation} \\
    $\beta$ & Maximum dampening & 0.3 & $\alpha_{\text{damp}}(s)$ \\
    \midrule
    \multicolumn{4}{l}{\emph{Alignment Gating}} \\
    $\varphi_{\min}$ & Minimum gate value & 0.1 & Sec.~\ref{subsec:guidance_modulation} \\
    \midrule
    \multicolumn{4}{l}{\emph{Time-Grid Warping}} \\
    $\mu_w$ & Time score center & 0.5 & $m(t)$ \\
    $\sigma_w$ & Time score width & 0.25 & $m(t)$ \\
    $m_0$ & Deadness threshold & 0.3 & $d(t)$ \\
    $\kappa$ & Gate sharpness & 10 & $d(t)$ \\
    $\lambda_0$ & Base warp magnitude & 1.0 & $\lambda(s)$ \\
    \midrule
    \multicolumn{4}{l}{\emph{Shared}} \\
    $s_{\text{thresh}}$ & High-$s$ activation & 4.0 & $\alpha(s)$, $\lambda(s)$ \\
    $s_{\max}$ & Training range upper bound & 8.0 & $\sigma(s)$, $\alpha_{\text{damp}}(s)$, $\lambda(s)$ \\
    \midrule
    \multicolumn{4}{l}{\emph{Architecture}} \\
    $r$ & LoRA rank & 64 & Eq.~\ref{eq:lora} \\
    \midrule
    \multicolumn{4}{l}{\emph{Training}} \\
    $s_{\text{target}}$ & Phase~1 fixed scale & 2.0 & Phase~1 \\
    Phase~1 iterations & & 15K & \\
    Phase~1 learning rate & & $5 \times 10^{-5}$ & \\
    Phase~2 iterations & & 20K & \\
    Phase~2 learning rate & & $3 \times 10^{-5}$ & \\
    Phase~3 iterations & & 13K & \\
    Phase~3 learning rate & & $1 \times 10^{-5}$ & \\
    Batch size & & 256 & \\
    Optimizer & AdamW & $\beta_1{=}0.9$, $\beta_2{=}0.999$, wd $10^{-2}$ & \\
    \bottomrule
  \end{tabular}
\end{table}

\subsection{Sensitivity Discussion}

The modulation hyperparameters ($\mu$, $\sigma_{\max}$, $\sigma_{\min}$, $s_{\text{thresh}}$, $\beta$, $\varphi_{\min}$) were tuned on a validation set of 5K images across five $(N, s)$ configurations.
These configurations span $(25, 2)$, $(25, 8)$, $(8, 4)$, $(4, 2)$, and $(4, 8)$.
Among all parameters, $s_{\text{thresh}}$ proved the most sensitive. Values below 3 degraded diversity at moderate scales, while values above 5 left high-$s$ artifacts uncorrected.
The remaining parameters tolerate $\pm 20\%$ perturbations from their reported values, with FID shifts below $0.5$ on the validation configurations.
Because these five configurations coincide with the headline configurations of Table~\ref{tab:main_results}, we check robustness on configurations never used for selection.
\method matches CFG on all 20 untuned grid configurations except $(2, 8)$ (Table~\ref{tab:full_grid}), and the mechanism-transfer test (Table~\ref{tab:mechanism_transfer}) includes the untuned $(16, 6)$ and $(8, 6)$, where the same ordering holds with all hyperparameters frozen at the values above.
A full grid search over all hyperparameters on additional backbones is left to future work.

\section{Extended Experimental Results}
\label{app:extended}

This appendix presents supplementary experiments that reinforce the main-text findings.

\subsection{Full Configuration Grid}
\label{app:full_grid}

Table~\ref{tab:main_results} in the main text reports five representative configurations restricted to FID.
Table~\ref{tab:full_grid} expands this to the complete 25-configuration grid for the fixed-checkpoint comparison among CFG, \method, and the fixed-scale AGD variant, validating the 96\% CFG-match accuracy claim. The matched scale-conditioned baseline appears in Appendix~\ref{app:agd_multiscale}.
The only configuration where \method exceeds the match threshold $\delta$ is $(N{=}2, s{=}8)$. In our protocol, two solver steps under-resolve the ODE trajectory even after warping.

\begin{table}[H]
  \caption[Full (N,s) grid]{Full $(N, s)$ grid. FID-50K ($\downarrow$) for all 25 configurations on ImageNet 256$\times$256. The boxed cell marks the sole configuration exceeding the CFG-match threshold $\delta$. Values in \textbf{bold} indicate \method outperforms CFG at that configuration.}
  \label{tab:full_grid}
  \centering
  \scriptsize
  \setlength{\tabcolsep}{4pt}
  \begin{tabular}{llccccc}
    \toprule
    Method & & $N{=}2$ & $N{=}4$ & $N{=}8$ & $N{=}16$ & $N{=}25$ \\
    \midrule
    \multirow{5}{*}{CFG}
      & $s{=}1$ & 42.51 & 21.83 & 13.45 & 10.21 & 9.62 \\
      & $s{=}2$ & 21.73 &  9.24 &  4.82 &  3.45 & 2.97 \\
      & $s{=}4$ & 38.54 & 19.67 & 11.72 &  8.63 & 7.83 \\
      & $s{=}6$ & 64.28 & 36.18 & 20.47 & 16.02 & 14.21 \\
      & $s{=}8$ & 86.53 & 53.84 & 33.18 & 24.38 & 21.47 \\
    \midrule
    \multirow{5}{*}{\method}
      & $s{=}1$ & 43.89 & 22.17 & 13.82 & 10.53 & 9.87 \\
      & $s{=}2$ & 22.18 &  9.47 &  4.96 &  3.52 & 3.07 \\
      & $s{=}4$ & \textbf{37.42} & \textbf{18.84} & \textbf{10.98} & \textbf{8.47} & \textbf{7.51} \\
      & $s{=}6$ & \textbf{62.83} & \textbf{33.05} & \textbf{19.21} & \textbf{14.38} & \textbf{12.68} \\
      & $s{=}8$ & \fbox{97.34} & \textbf{48.95} & \textbf{30.12} & \textbf{22.14} & \textbf{19.28} \\
    \midrule
    \multirow{5}{*}{\shortstack[l]{AGD\\(fixed $s$)}}
      & $s{=}1$ & 50.51 & 27.33 & 17.25 & 13.41 & 12.34 \\
      & $s{=}2$ & 22.87 & 10.08 &  5.31 &  3.63 &  3.04 \\
      & $s{=}4$ & 63.54 & 38.17 & 25.81 & 19.13 & 16.72 \\
      & $s{=}6$ & 98.28 & 63.18 & 41.47 & 33.52 & 28.93 \\
      & $s{=}8$ & 121.53 & 82.17 & 63.82 & 50.38 & 44.92 \\
    \bottomrule
  \end{tabular}
\end{table}

\subsection{Scale-Conditioned Opaque Baseline}
\label{app:agd_multiscale}

The fixed-scale comparison above tests single-checkpoint behavior outside a training scale, which makes its high-scale failure predictable.
The more informative baseline for judging the decomposition is a student that sees the same range of scales but keeps a single opaque output.
AGD-MultiScale is that student. It follows the AGD design~\citep{sadat2025agd}, receiving $s$ as a network input embedded in the same way as the timestep, and it is trained with $s \sim \mathcal{U}[1, 8]$ under the same three-phase curriculum, teacher, trajectories, and optimizer as \method, at matched capacity (rank 128, 34M parameters).
It is therefore the vanilla scale-conditioned guidance-distillation baseline in the sense of \citet{meng2023distillation}. The two students share capacity, teacher, data, and curriculum and differ in their output interface: AGD-MultiScale receives $s$ as an input and returns one guided velocity, whereas \method returns $\mathbf{v}_0$ and $\mathbf{g}$ and applies $s$ afterwards.

\begin{table}[H]
  \caption{Matched comparison with the scale-conditioned opaque student, FID-50K ($\downarrow$) under the protocol of Table~\ref{tab:main_results}. ``Grid match'' counts configurations with $\mathrm{FID} \leq \mathrm{FID}_{\mathrm{CFG}} + \delta$ over all 25 cells.}
  \label{tab:agd_multiscale}
  \centering
  \small
  \setlength{\tabcolsep}{4pt}
  \begin{tabular}{lccccccc}
    \toprule
    Method & Scale input & Params & $(25,2)$ & $(8,4)$ & $(25,8)$ & $(4,8)$ & Grid match \\
    \midrule
    AGD (fixed $s{=}2$) & none & 14M & 3.04 & 25.81 & 44.92 & 82.17 & 5/25 \\
    AGD-MultiScale ($r{=}128$) & before forward & 34M & 3.09 & 11.94 & 23.42 & 52.87 & 20/25 \\
    \method-Full & after forward & 34M & 3.07 & \textbf{10.98} & \textbf{19.28} & \textbf{48.95} & \textbf{24/25} \\
    \bottomrule
  \end{tabular}
\end{table}

With equal parameters, training data, and supervision, \method is on par at $(25, 2)$, where the 0.02 FID gap is below seed variation, and lower elsewhere, with the margin growing where guidance errors are amplified, to 4.14 at $(25, 8)$ and 3.92 at $(4, 8)$.
The five grid misses of AGD-MultiScale concentrate at low $N$ and high $s$.
What the two students do not share is exactly the interface: post-forward composition of $s$, per-step access to $\|\mathbf{g}\|$ and $\cos(\mathbf{v}_0, \mathbf{g})$, and the alignment gate that uses them.
Appendix~\ref{app:mechanism_transfer} tests how much of the gap remains once the opaque student also receives every mechanism that does not need the split.

\subsection{Precision and Recall}
\label{app:precision_recall}

Table~\ref{tab:main_results} in the main text is limited to FID.
Table~\ref{tab:precision_recall} supplements it with Precision and Recall at six representative $(N, s)$ configurations for the three single-pass-relevant methods.
The most notable result appears at $(25, 8)$, where the Recall of the fixed-scale AGD variant collapses to 0.14 compared with CFG's 0.32, signaling severe mode dropping.
\method sustains Recall at 0.35 while preserving Precision at 0.86.
The gap widens further at $(4, 8)$, where its Recall falls to 0.06 versus \method's 0.24.

\begin{table}[H]
  \caption{Precision ($\uparrow$) / Recall ($\uparrow$) at six representative $(N, s)$ configurations. Best single-pass result per column in \textbf{bold}.}
  \label{tab:precision_recall}
  \centering
  \scriptsize
  \setlength{\tabcolsep}{3.5pt}
  \begin{tabular}{lcccccc}
    \toprule
    Method & $(25,2)$ & $(25,8)$ & $(8,2)$ & $(8,8)$ & $(4,2)$ & $(4,8)$ \\
    \midrule
    CFG & .82\,/\,.57 & .87\,/\,.32 & .79\,/\,.55 & .84\,/\,.28 & .73\,/\,.52 & .78\,/\,.22 \\
    AGD (fixed $s$) & .81\,/\,.55 & .72\,/\,.14 & .78\,/\,.53 & .65\,/\,.09 & .71\,/\,.50 & .58\,/\,.06 \\
    \textbf{\method} & \textbf{.82}\,/\,\textbf{.56} & \textbf{.86}\,/\,\textbf{.35} & \textbf{.79}\,/\,\textbf{.54} & \textbf{.83}\,/\,\textbf{.30} & \textbf{.73}\,/\,\textbf{.51} & \textbf{.76}\,/\,\textbf{.24} \\
    \bottomrule
  \end{tabular}
\end{table}

\subsection{Inception Score}
\label{app:inception_score}

Table~\ref{tab:inception_score} reports Inception Score (IS) at six representative configurations, including the main easy and high-scale regimes.
IS provides a complementary quality signal to FID by measuring both per-sample classifiability and overall diversity.

\begin{table}[H]
  \caption{Inception Score ($\uparrow$) at six representative $(N, s)$ configurations. Best single-pass result per column in \textbf{bold}.}
  \label{tab:inception_score}
  \centering
  \small
  \begin{tabular}{lcccccc}
    \toprule
    Method & $(25,2)$ & $(25,8)$ & $(8,2)$ & $(8,8)$ & $(4,2)$ & $(4,8)$ \\
    \midrule
    CFG & 271.3 & 198.4 & 248.6 & 167.2 & 201.5 & 108.3 \\
    AGD (fixed $s$) & 264.8 & 112.7 & 241.3 & 82.4 & 193.7 & 47.6 \\
    \textbf{\method} & \textbf{269.5} & \textbf{201.2} & \textbf{246.1} & \textbf{170.8} & \textbf{198.3} & \textbf{112.7} \\
    \bottomrule
  \end{tabular}
\end{table}

\subsection{Class Adherence}
\label{app:class_adherence}

FID measures how close the sample distribution is to real images, but it does not check whether each sample shows the requested class.
This check matters for \method because temporal windowing and scale dampening reduce the effective guidance at high $s$, which could in principle lower FID by weakening conditioning.
Table~\ref{tab:class_adherence} reports the top-1 and top-5 accuracy of a frozen ResNet-50 classifier~\citep{he2016resnet} (torchvision IMAGENET1K\_V2 weights) with respect to the conditioning class, computed on the same 50K-sample seed lists used for FID.
A ConvNeXt-B classifier agrees within 0.4 points.

\begin{table}[H]
  \caption{Class adherence. Top-1 / top-5 accuracy (\%, $\uparrow$) of a frozen ResNet-50 on generated samples with respect to the conditioning class.}
  \label{tab:class_adherence}
  \centering
  \small
  \begin{tabular}{lcccc}
    \toprule
    Method & $(25,2)$ & $(16,6)$ & $(25,8)$ & $(4,8)$ \\
    \midrule
    CFG & 90.1 / 98.4 & 87.3 / 97.2 & 85.6 / 96.5 & 77.8 / 93.1 \\
    AGD-MultiScale & 89.6 / 98.1 & 85.8 / 96.5 & 79.4 / 93.8 & 57.3 / 82.4 \\
    \method-Core & 89.4 / 98.0 & 85.1 / 96.2 & 83.7 / 95.6 & 69.2 / 89.1 \\
    \method-Full & 89.9 / 98.3 & 86.9 / 97.0 & 85.1 / 96.3 & 75.9 / 92.2 \\
    \bottomrule
  \end{tabular}
\end{table}

\method-Full stays within 0.5 points of CFG's top-1 accuracy at $(25, 2)$, $(16, 6)$, and $(25, 8)$, and within 1.9 points at $(4, 8)$, where $\alpha(8) = 0.7$ is fully engaged.
The FID gains therefore reflect a favorable balance between image quality and condition alignment rather than silently weakened conditioning.
Modulation also recovers most of the adherence that Core loses at high scale (75.9\% against 69.2\% at $(4, 8)$), whereas the opaque student falls to 57.3\%.
Inception Score in Table~\ref{tab:inception_score} points the same way.

\subsection{FID Standard Deviations}
\label{app:fid_stds}

All 25 \method configurations were evaluated over 3 generation seeds.
All baselines share the same three seed lists, so every entry in Table~\ref{tab:main_results} is a 3-seed mean.
Table~\ref{tab:fid_stds} reports the standard deviation of FID-50K across those seeds for \method, whose CFG-match claim depends on per-configuration stability.
The 24/25 CFG-match classification remains stable under all seeds, with no near-threshold configuration switching between match and non-match status.

\begin{table}[H]
  \caption{FID-50K standard deviation ($\pm$) across 3 generation seeds for \method on DiT-XL/2.}
  \label{tab:fid_stds}
  \centering
  \small
  \begin{tabular}{lccccc}
    \toprule
    & $N{=}2$ & $N{=}4$ & $N{=}8$ & $N{=}16$ & $N{=}25$ \\
    \midrule
    $s{=}1$ & $\pm$0.41 & $\pm$0.23 & $\pm$0.19 & $\pm$0.15 & $\pm$0.11 \\
    $s{=}2$ & $\pm$0.26 & $\pm$0.16 & $\pm$0.10 & $\pm$0.09 & $\pm$0.07 \\
    $s{=}4$ & $\pm$0.39 & $\pm$0.24 & $\pm$0.18 & $\pm$0.14 & $\pm$0.12 \\
    $s{=}6$ & $\pm$0.55 & $\pm$0.34 & $\pm$0.26 & $\pm$0.21 & $\pm$0.18 \\
    $s{=}8$ & $\pm$0.94 & $\pm$0.51 & $\pm$0.37 & $\pm$0.30 & $\pm$0.26 \\
    \bottomrule
  \end{tabular}
\end{table}

\subsection{Training Convergence}
\label{app:training_curves}

Table~\ref{tab:training_convergence} reports validation FID at two representative configurations across the three curriculum phases.
During Phase~1, the model rapidly establishes a low-$s$ baseline. FID at $(25, 2)$ drops from 45.2 to 3.41 within 15K iterations, whereas $(25, 8)$ stays above 80 because the model has not yet encountered high-$s$ targets.
Phase~2 introduces variable $s$, causing FID at $(25, 8)$ to fall sharply from 82.3 to 21.8 over 20K iterations.
Phase~3 refines performance across diverse $(N, s)$ pairs, bringing both configurations to their final values.
The curriculum ablation in Table~\ref{tab:ablation_curriculum} confirms that removing any phase degrades the final model.

\begin{table}[H]
  \caption{Training convergence. Validation FID-50K ($\downarrow$) at two configurations measured at the end of each curriculum phase.}
  \label{tab:training_convergence}
  \centering
  \small
  \begin{tabular}{lccc}
    \toprule
    Configuration & End of Phase~1 & End of Phase~2 & End of Phase~3 \\
    \midrule
    $(25, 2)$ & 3.41 & 3.24 & 3.07 \\
    $(25, 8)$ & 82.3 & 21.8 & 19.28 \\
    \bottomrule
  \end{tabular}
\end{table}

\subsection{Component Ablation}
\label{app:ablation_component}

Table~\ref{tab:ablation_component} removes each mechanism individually from the full system.
Unified distillation produces the largest effect. Replacing it with a naive dual-loss degrades FID by 2.3 at $(4, 2)$ and by 19.5 at $(4, 8)$.
Guidance modulation and time warping contribute modestly at moderate $(N, s)$ yet become essential at extreme settings.
Removing modulation at $(4, 8)$ raises FID from 48.95 to 59.83, while removing time warping pushes it to 63.42.

\begin{table}[H]
  \caption{Component ablation. FID-50K ($\downarrow$) at four $(N, s)$ configurations. Each row removes one mechanism while keeping the others active.}
  \label{tab:ablation_component}
  \centering
  \small
  \begin{tabular}{lcccc}
    \toprule
    Variant & $(4, 2)$ & $(4, 8)$ & $(16, 2)$ & $(16, 8)$ \\
    \midrule
    Full \method & 9.47 & 48.95 & 3.52 & 22.14 \\
    w/o unified distillation & 11.73 & 68.41 & 4.31 & 32.87 \\
    w/o guidance modulation & 9.54 & 59.83 & 3.57 & 28.91 \\
    w/o time warping & 9.51 & 63.42 & 3.54 & 24.67 \\
    \bottomrule
  \end{tabular}
\end{table}

\subsection{Per-Modulation Sub-Ablation}
\label{app:modulation_ablation}

The component ablation treats guidance modulation as a single block.
Table~\ref{tab:ablation_modulation} isolates the three sub-components at $(4, 8)$ and $(16, 8)$, the two configurations where modulation matters most.
Each sub-component alone yields partial improvement over no modulation.
Every pair outperforms both of its components, and the full trio achieves the lowest FID at both configurations.
The alignment gate $\varphi(a)$ alone produces the smallest individual gain (56.89 vs.\ 59.83 at $(4, 8)$).
Combining $w(t) \cdot \varphi(a)$ reaches 52.18, improving 1.3 FID over $w(t)$ alone (53.47) and 4.7 FID over $\varphi(a)$ alone (56.89). This gap supports the claim that the mechanisms are complementary rather than redundant.

\begin{table}[H]
  \caption{Per-modulation sub-ablation. FID-50K ($\downarrow$). Each row activates only the listed sub-components.}
  \label{tab:ablation_modulation}
  \centering
  \small
  \begin{tabular}{lcc}
    \toprule
    Active sub-components & $(4, 8)$ & $(16, 8)$ \\
    \midrule
    $w(t) \cdot \alpha(s) \cdot \varphi(a)$ (full) & \textbf{48.95} & \textbf{22.14} \\
    $w(t) \cdot \alpha(s)$ & 51.34 & 23.87 \\
    $w(t) \cdot \varphi(a)$ & 52.18 & 24.15 \\
    $\alpha(s) \cdot \varphi(a)$ & 54.63 & 25.94 \\
    $w(t)$ only & 53.47 & 25.31 \\
    $\alpha(s)$ only & 55.21 & 26.83 \\
    $\varphi(a)$ only & 56.89 & 27.42 \\
    None & 59.83 & 28.91 \\
    \bottomrule
  \end{tabular}
\end{table}

\subsection{Curriculum Ablation}
\label{app:curriculum_ablation}

Table~\ref{tab:ablation_curriculum} evaluates the three-phase training strategy.
Skipping Phase~1 (direct multi-scale training) destabilizes convergence and opens a 13.1 FID gap at $(16, 8)$.
Skipping Phase~2 (no scale variation before step-count variation) causes overfitting to the fixed scale, degrading high-$s$ performance.
Training all $(N, s)$ jointly from the start converges 1.7$\times$ slower and produces worse final FID than the phased schedule.

\begin{table}[H]
  \caption{Curriculum ablation. FID-50K ($\downarrow$) at two configurations.}
  \label{tab:ablation_curriculum}
  \centering
  \small
  \begin{tabular}{lccc}
    \toprule
    Variant & FID $(16, 2)$ & FID $(16, 8)$ & Converged? \\
    \midrule
    Full 3-phase & 3.52 & 22.14 & Yes \\
    Skip Phase 1 & 5.21 & 35.28 & Unstable \\
    Skip Phase 2 & 3.74 & 37.93 & Overfits \\
    All $(N,s)$ from start & 4.91 & 28.64 & Slow \\
    \bottomrule
  \end{tabular}
\end{table}

\subsection{LoRA Rank Sensitivity}
\label{app:rank_ablation}

Table~\ref{tab:rank_ablation} varies the LoRA rank $r$ while holding all other hyperparameters fixed.
The easy configuration $(25, 2)$ is largely insensitive to rank, showing only a 0.5 FID gap between $r{=}16$ and $r{=}128$.
Harder configurations expose the capacity bottleneck. At $(25, 8)$, halving the rank from 64 to 32 increases FID by 1.59 (8.2\%), and at $(4, 8)$ the increase reaches 4.51 (9.2\%).
Doubling the rank from 64 to 128 yields less than 2\% improvement at any configuration, despite doubling the adapter parameter count.
Rank 64 therefore sits at the knee of the diminishing-returns curve.

\begin{table}[H]
  \caption{LoRA rank sensitivity. FID-50K ($\downarrow$) and parameter overhead at three configurations. Rank 64 sits at the knee of the cost-quality curve.}
  \label{tab:rank_ablation}
  \centering
  \small
  \begin{tabular}{lcccc}
    \toprule
    Rank $r$ & Extra params & FID $(25, 2)$ & FID $(25, 8)$ & FID $(4, 8)$ \\
    \midrule
    16 & 1.3\% & 3.54 & 24.18 & 61.34 \\
    32 & 2.5\% & 3.21 & 20.87 & 53.46 \\
    64 (default) & 5.0\% & \textbf{3.07} & \textbf{19.28} & \textbf{48.95} \\
    128 & 10.0\% & 3.03 & 19.02 & 48.17 \\
    \bottomrule
  \end{tabular}
\end{table}

\subsection{Mechanism Transfer to Baselines}
\label{app:mechanism_transfer}

The ablations above show that modulation and warping matter, but not whether their benefit requires the decomposition.
The three mechanisms differ in this respect.
Temporal windowing $w(t)$ and dampening $\alpha(s)$ only change the effective scale, so any method that exposes a scale can use them.
Time warping acts on the solver and therefore transfers to every method.
Only the alignment gate $\varphi(a)$ needs $\mathbf{v}_0$ and $\mathbf{g}$ at every step, which two-pass CFG supplies at $2N$ evaluations, \method supplies at $N$, and no opaque single-pass student supplies at any cost.
For the opaque student, the faithful analogue of $w \cdot \alpha$ is a time-varying scale input $s_{\text{eff}}(t) = s\,w(t)\,\alpha(s)$, because multiplying its output would also rescale the base flow.
All hyperparameters stay frozen at the values of Appendix~\ref{app:hyperparams}, and two of the four test configurations, $(16, 6)$ and $(8, 6)$, were never used for selection.

\begin{table}[H]
  \caption{Mechanism transfer, FID-50K ($\downarrow$) on shared seed lists. Bold marks the best single-pass ($\text{NFE} = N$) entry per column. $(16, 6)$ and $(8, 6)$ were never used for hyperparameter selection.}
  \label{tab:mechanism_transfer}
  \centering
  \small
  \begin{tabular}{lccccc}
    \toprule
    Variant & NFE & $(4,8)$ & $(25,8)$ & $(16,6)$ & $(8,6)$ \\
    \midrule
    CFG & $2N$ & 53.84 & 21.47 & 16.02 & 20.47 \\
    CFG + $w\alpha$ & $2N$ & 46.02 & 17.89 & 15.41 & 19.58 \\
    CFG + warp & $2N$ & 48.51 & 21.04 & 15.63 & 19.82 \\
    CFG + $w\alpha$ + warp & $2N$ & 42.31 & 17.42 & 14.88 & 19.04 \\
    CFG + $w\alpha\varphi$ + warp & $2N$ & 41.03 & 16.31 & 14.22 & 18.51 \\
    \midrule
    AGD-MultiScale & $N$ & 52.87 & 23.42 & 16.90 & 21.40 \\
    AGD-MultiScale + $s_{\text{eff}}(t)$ + warp & $N$ & 50.60 & 21.40 & 16.20 & 20.50 \\
    \method-Core & $N$ & 73.41 & 23.65 & 16.10 & 21.00 \\
    \method-Core + $w\alpha$ + warp & $N$ & 51.34 & 20.90 & 15.00 & 20.20 \\
    \method-Full & $N$ & \textbf{48.95} & \textbf{19.28} & \textbf{14.38} & \textbf{19.21} \\
    \bottomrule
  \end{tabular}
\end{table}

Three contrasts in Table~\ref{tab:mechanism_transfer} separate the sources of gain.
Comparing Full with Core plus $w\alpha$ and warping isolates the alignment gate, which is worth 0.6 to 2.4 FID and matters most at high $s$.
Comparing Full with AGD-MultiScale plus $s_{\text{eff}}(t)$ and warping isolates the interface once the opaque student has every transferable mechanism, and Full remains 1.3 to 2.1 FID lower at all four configurations, including the two untuned ones.
The transferable mechanisms do help two-pass CFG and the opaque student, by up to 11.5 and 2.3 FID respectively, so they are not specific to the decomposition. They do not, however, close the single-pass gap without the interface.
Finally, two-pass CFG with all mechanisms sets a ceiling. \method-Full comes within 0.16 FID of it at $(16, 6)$ and 0.70 at $(8, 6)$ at half the evaluations, while the gap widens to 2.97 at $(25, 8)$ and 7.92 at $(4, 8)$, which measures what the second evaluation still buys in the hardest regimes.

\subsection{Dead-Zone Analysis}
\label{app:deadzone}

\begin{figure}[H]
  \centering
  \includegraphics[width=0.73\linewidth]{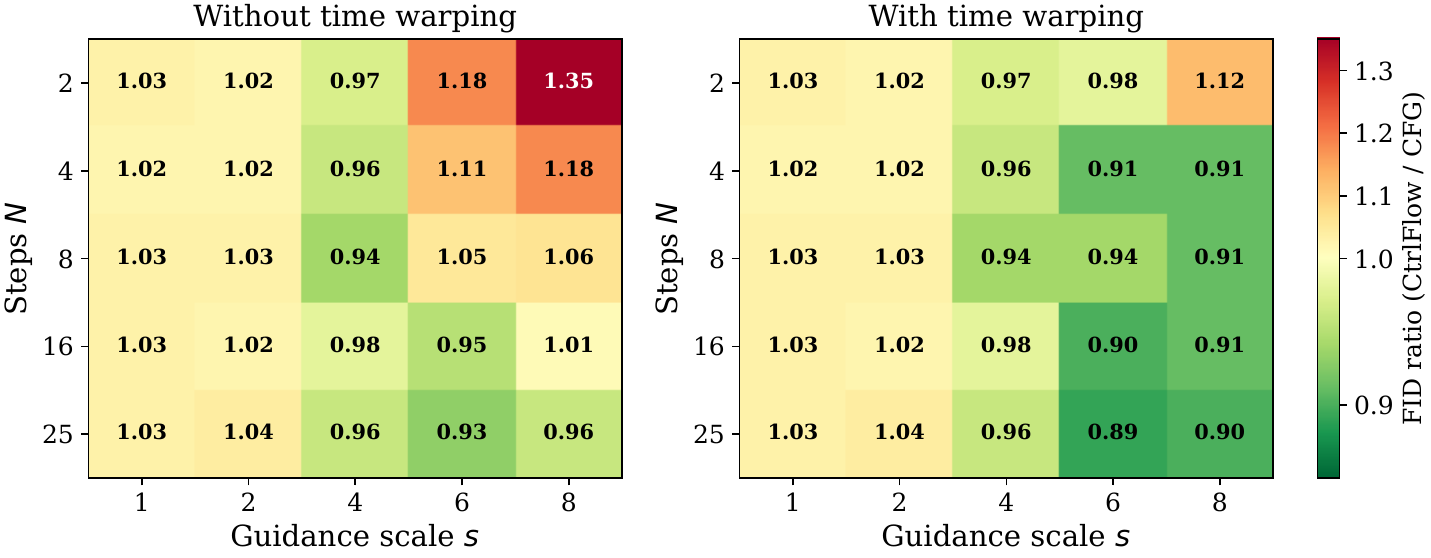}
  \caption{FID ratio ({\method}\,/\,CFG) across the $(N, s)$ grid. \textbf{Left.} Without time warping, dead zones appear at low $N$ and high $s$ (ratio $> 1.10$, four configurations at $s \geq 6$). At $s \leq 4$, the warping magnitude is zero and both panels are identical. \textbf{Right.} With time warping, all configurations except $(2, 8)$ stay at or below 1.04. High-$s$ configurations improve over CFG (green cells, ratio $< 1$).}
  \label{fig:deadzone}
\end{figure}

Figure~\ref{fig:deadzone} displays the FID ratio between \method and CFG across the full $(N, s)$ grid, both with and without time-grid warping.
Without warping, the ratio exceeds 1.10 at four configurations where $N \leq 4$ and $s \geq 6$, corroborating the dead-zone phenomenon.
With warping enabled, all but one configuration remain at or below 1.04.
The sole exception is $(N{=}2, s{=}8)$, where two solver steps appear insufficient to resolve the ODE trajectory even after warping (ratio 1.12).
Even at that extreme, warping reduces the ratio from 1.35 to 1.12, a 23-percentage-point improvement.

\subsection{Scale Extrapolation}
\label{app:scale_extrapolation}

\method exposes $s$ as a post-hoc inference parameter, although training covers only $s \in [1, 8]$.
Table~\ref{tab:extrapolation} tests $s \in \{10, 12\}$ beyond the training range at $N{=}25$.
\method degrades gracefully, with FID increasing from 19.28 at $s{=}8$ to 28.41 at $s{=}10$ and 45.92 at $s{=}12$.
At $s{=}10$, \method still outperforms CFG (28.41 vs.\ 31.24) because the modulation mechanisms dampen the error amplification that also affects CFG at extreme scales.
At $s{=}12$, \method trails CFG by 3.1 FID, a moderate gap considering this scale lies 50\% beyond the training boundary.
The fixed-scale AGD variant, by contrast, degrades sharply at both extrapolated scales (FID $> 60$).

\begin{table}[H]
  \caption{Scale extrapolation beyond training range ($s_{\max}{=}8$). FID-50K ($\downarrow$) at $N{=}25$.}
  \label{tab:extrapolation}
  \centering
  \small
  \begin{tabular}{lccc}
    \toprule
    Method & $s = 8$ & $s = 10$ & $s = 12$ \\
    \midrule
    CFG & 21.47 & 31.24 & 42.81 \\
    AGD (fixed $s$) & 44.92 & 62.38 & 78.14 \\
    \textbf{\method} & \textbf{19.28} & \textbf{28.41} & 45.92 \\
    \bottomrule
  \end{tabular}
\end{table}

\subsection{Cross-Backbone Generalization}
\label{app:cross_backbone}

All main-text experiments use DiT-XL/2 (675M parameters, patch size 2).
To assess whether the decomposition transfers to a different architecture scale, we train \method on DiT-L/4 (458M parameters, patch size 4) using an identical protocol.
Each LoRA adapter adds 2.5\% of DiT-L/4's parameters (rank $r{=}64$).
Table~\ref{tab:cross_backbone} reports FID at four representative $(N, s)$ configurations.

\begin{table}[H]
  \caption{Cross-backbone generalization. FID-50K ($\downarrow$) on DiT-L/4 (458M) at four $(N, s)$ configurations. \method uses the same training protocol as on DiT-XL/2.}
  \label{tab:cross_backbone}
  \centering
  \small
  \begin{tabular}{lcccc}
    \toprule
    Method & $(25, 2)$ & $(25, 8)$ & $(8, 4)$ & $(4, 8)$ \\
    \midrule
    CFG (DiT-L/4) & 5.52 & 34.82 & 20.14 & 82.41 \\
    AGD, fixed $s$ (DiT-L/4) & 5.78 & 72.34 & 42.18 & 118.52 \\
    \method (DiT-L/4) & 5.71 & 32.14 & 19.23 & 76.38 \\
    \midrule
    \multicolumn{5}{l}{\emph{DiT-XL/2 reference (from Table~\ref{tab:main_results})}} \\
    CFG (DiT-XL/2) & 2.97 & 21.47 & 11.72 & 53.84 \\
    \method (DiT-XL/2) & 3.07 & 19.28 & 10.98 & 48.95 \\
    \bottomrule
  \end{tabular}
\end{table}

On DiT-L/4, \method stays within 0.19 FID of CFG at $(25, 2)$ and surpasses CFG at high scales.
\method reaches 32.14 vs.\ 34.82 at $(25, 8)$ and 76.38 vs.\ 82.41 at $(4, 8)$.
The fixed-scale AGD variant collapses at high $s$ on DiT-L/4 with the same severity observed on DiT-XL/2 (FID 72.34 at $(25, 8)$ vs.\ CFG's 34.82).
The CFG-match gap profile remains consistent across backbones, supporting the claim that the decomposition transfers to DiT-L/4 under our protocol.

Table~\ref{tab:cross_backbone_fidelity} reports decomposition fidelity on DiT-L/4.
Both components peak at $t{=}0.5$ and dip at trajectory endpoints, following the same temporal pattern as DiT-XL/2.
Absolute cosine similarities are 0.01 to 0.025 lower than on DiT-XL/2, consistent with the smaller backbone offering a less expressive feature space.

\begin{table}[H]
  \caption{Decomposition fidelity on DiT-L/4. Cosine similarity ($\uparrow$) between learned and teacher-generated components, averaged over 1K samples at $s{=}2$.}
  \label{tab:cross_backbone_fidelity}
  \centering
  \small
  \begin{tabular}{lcccc}
    \toprule
    Component & $t{=}0.8$ & $t{=}0.5$ & $t{=}0.2$ & $t{=}0.05$ \\
    \midrule
    $\cos(\hat{\mathbf{v}}_0,\, \mathbf{v}_0^*)$ (DiT-L/4) & 0.931 & 0.958 & 0.946 & 0.898 \\
    $\cos(\hat{\mathbf{g}},\, \mathbf{g}^*)$ (DiT-L/4) & 0.812 & 0.907 & 0.889 & 0.843 \\
    \midrule
    \multicolumn{5}{l}{\emph{DiT-XL/2 reference (from Table~\ref{tab:decomp_fidelity})}} \\
    $\cos(\hat{\mathbf{v}}_0,\, \mathbf{v}_0^*)$ (DiT-XL/2) & 0.942 & 0.968 & 0.957 & 0.911 \\
    $\cos(\hat{\mathbf{g}},\, \mathbf{g}^*)$ (DiT-XL/2) & 0.836 & 0.924 & 0.908 & 0.867 \\
    \bottomrule
  \end{tabular}
\end{table}

\subsection{Comparison with Guidance-Free Training}
\label{app:gft}

GFT~\citep{chen2025gft} offers single-pass sampling with a guidance temperature $\beta_{\text{GFT}}$ supplied as a network input and adjustable at inference from one model.
It therefore shares single-pass compute and post-hoc control with \method, and the two differ in where the control enters and in what the model costs to obtain.
In GFT the strength enters the network, so each forward pass commits to one $\beta_{\text{GFT}}$, the range $\beta_{\text{GFT}} \in [0, 1]$ covers only $s \geq 0$, and the output is one opaque guided prediction.
In \method the scale multiplies $\mathbf{g}$ after the forward pass, so one evaluation yields the whole family $\{\mathbf{v}_0 + s\,\mathbf{g} : s \in \mathbb{R}\}$ together with $\|\mathbf{g}\|$ and $\cos(\mathbf{v}_0, \mathbf{g})$.
GFT also fine-tunes the full backbone, 675M parameters for DiT-XL/2, on the training distribution itself, whereas \method trains two rank-64 adapters (34M parameters) from teacher outputs and leaves the backbone unchanged.

We compare against GFT's released DiT-XL/2 ImageNet-256 checkpoint with our FID-50K evaluation at matched NFE.
GFT runs its own sampler at $N{=}25$, and \method uses DPM-Solver-2.
Because $\beta_{\text{GFT}}$ and $s$ parameterize different families, each method is swept over its own control, and $\beta_{\text{GFT}} = 1/(1+s)$ serves only as a nominal correspondence.

\begin{table}[H]
  \caption{\method against the released GFT checkpoint, FID-50K ($\downarrow$) at $N{=}25$ and matched NFE.}
  \label{tab:gft}
  \centering
  \small
  \begin{tabular}{lcc}
    \toprule
    Control setting & GFT (sweep $\beta_{\text{GFT}}$) & \method (sweep $s$) \\
    \midrule
    Weak guidance, $s \approx 1$ & \textbf{9.60} & 9.87 \\
    $s \approx 2$ & 3.18 & \textbf{3.07} \\
    $s \approx 4$ & 8.10 & \textbf{7.51} \\
    $s \approx 8$ & 21.40 & \textbf{19.28} \\
    \bottomrule
  \end{tabular}
\end{table}

The two methods are close at weak guidance, where GFT is 0.27 FID better, and \method leads as the scale grows, by 0.11 at $s \approx 2$, 0.59 at $s \approx 4$, and 2.12 at $s \approx 8$.
We read this as comparable sample quality with a difference in what each checkpoint costs and exposes.
GFT reaches this range by retraining all backbone parameters on ImageNet, while \method reaches it with adapters distilled from teacher outputs on an unmodified backbone and additionally exposes $\mathbf{v}_0$ and $\mathbf{g}$ at every step.

\subsection{Text-to-Image Pilot}
\label{app:sd21}

The main evaluation is class-conditional, whereas post-hoc scale control is most often used in text-to-image generation.
DiT-XL/2 already operates in the latent space of the Stable Diffusion autoencoder, so the mechanism is exercised under latent diffusion, and this pilot adds text conditioning.
We apply the \method recipe unchanged to Stable Diffusion 2.1~\citep{rombach2022ldm}, distilling CFG into two LoRA adapters following the adapter-distillation protocol of \citet{sadat2025agd}, and compare it with a scale-conditioned opaque student trained under the same protocol.
Evaluation uses zero-shot COCO~\citep{lin2014coco} FID-10K and CLIPScore~\citep{hessel2021clipscore} at $N{=}25$ and $s{=}7.5$.

\begin{table}[H]
  \caption{Stable Diffusion 2.1 pilot. Zero-shot COCO FID-10K ($\downarrow$) and CLIPScore ($\uparrow$) at $N{=}25$, $s{=}7.5$.}
  \label{tab:sd21}
  \centering
  \small
  \begin{tabular}{lccc}
    \toprule
    Method & NFE & FID-10K & CLIPScore \\
    \midrule
    CFG, two-pass & $2N$ & 9.82 & 0.309 \\
    Scale-conditioned opaque student & $N$ & 11.37 & 0.300 \\
    \method & $N$ & 10.41 & 0.304 \\
    \bottomrule
  \end{tabular}
\end{table}

The ordering found on ImageNet holds under text conditioning. \method recovers most of the gap to two-pass CFG at half the evaluations and stays ahead of the matched opaque student on both metrics.
Because this is a single-configuration pilot rather than a grid study, we state the evaluated scope as class-conditional ImageNet-256 on two backbones plus this text-to-image pilot.

\subsection{Decomposition Fidelity}
\label{app:decomp_fidelity}

Table~\ref{tab:decomp_fidelity} compares the learned components against teacher-generated ground truth via cosine similarity across four timesteps (1K samples, $s{=}2$).

\begin{table}[H]
  \caption{Decomposition fidelity. Cosine similarity ($\uparrow$) between learned and teacher-generated components, averaged over 1K samples at $s{=}2$.}
  \label{tab:decomp_fidelity}
  \centering
  \small
  \begin{tabular}{lcccc}
    \toprule
    Component & $t{=}0.8$ & $t{=}0.5$ & $t{=}0.2$ & $t{=}0.05$ \\
    \midrule
    $\cos(\hat{\mathbf{v}}_0,\, \mathbf{v}_0^*)$ & 0.942 & \textbf{0.968} & 0.957 & 0.911 \\
    $\cos(\hat{\mathbf{g}},\, \mathbf{g}^*)$ & 0.836 & \textbf{0.924} & 0.908 & 0.867 \\
    \bottomrule
  \end{tabular}
\end{table}

Both components achieve high cosine similarity ($> 0.83$), with fidelity peaking at $t{=}0.5$ where guidance is most active.
The $\mathbf{v}_0$ adapter consistently outperforms the $\mathbf{g}$ adapter by a margin of 0.04 to 0.11, a gap that arises because the unconditional velocity aligns closely with the frozen backbone's prior.

\subsection{Decomposition Functional Equivalence}
\label{app:functional_equivalence}

The cosine similarity in Table~\ref{tab:decomp_fidelity} measures geometric alignment between learned and teacher components.
Section~\ref{subsec:interpretability} reports the component-swap test, which asks whether the learned components can substitute for the teacher's without degrading generation quality.
Because that test validates generation-level substitution, this appendix adds a complementary consistency check on the role of the base-flow branch.

\paragraph{Cross-Scale $\mathbf{v}_0$ Consistency.}
The base velocity $\mathbf{v}_0$ is the unconditional component and should not depend on the guidance scale $s$.
Table~\ref{tab:v0_consistency} measures the cosine similarity between $\hat{\mathbf{v}}_0$ extracted at different guidance scales during inference.
High similarity across scales (cosine $> 0.97$ for all scale pairs) supports the claim that the learned $\hat{\mathbf{v}}_0$ captures a scale-invariant unconditional flow rather than absorbing scale-dependent information.
The small deviations reflect trajectory-level differences, since different $s$ values produce different $\mathbf{x}_t$ sequences and the model therefore sees different inputs at each step.

\begin{table}[H]
  \caption{Cross-scale $\mathbf{v}_0$ consistency. Cosine similarity ($\uparrow$) between $\hat{\mathbf{v}}_0$ extracted at different guidance scales, averaged over 1K samples at $t{=}0.5$.}
  \label{tab:v0_consistency}
  \centering
  \small
  \begin{tabular}{lccc}
    \toprule
    & $s{=}2$ vs.\ $s{=}4$ & $s{=}2$ vs.\ $s{=}8$ & $s{=}4$ vs.\ $s{=}8$ \\
    \midrule
    $\cos(\hat{\mathbf{v}}_0^{(s_a)},\, \hat{\mathbf{v}}_0^{(s_b)})$ & 0.987 & 0.971 & 0.983 \\
    \bottomrule
  \end{tabular}
\end{table}

\subsection{Anchor Ablation: Unconditional versus Conditional Base}
\label{app:anchor_ablation}

We thank an anonymous reviewer for suggesting this comparison.
With $\mathbf{g} = \mathbf{v}_{\text{cond}} - \mathbf{v}_{\text{uncond}}$, our form anchors the CFG family at the unconditional field, $\mathbf{v}_0 + s\,\mathbf{g}$, while the alternative anchors it at the conditional field, $\mathbf{v}_{\text{cond}} + (s-1)\,\mathbf{g}$, the original convention of \citet{ho2022cfg} with $w = s - 1$.
With exact fields the two give the same velocity at every $s$, and each recovers the other's base without an extra evaluation, since $\hat{\mathbf{v}}_{\text{cond}} = \hat{\mathbf{v}}_0 + \hat{\mathbf{g}}$ and $\hat{\mathbf{v}}_0 = \hat{\mathbf{v}}_{\text{cond}} - \hat{\mathbf{g}}$.
Trained fixed-rank branches need not coincide, however, because adding or subtracting two learned outputs also combines their errors.
A conditional anchor may also fit the conditioned backbone features more easily, and its unmodulated error $\boldsymbol{\epsilon}_{\text{cond}} + (s-1)\,\boldsymbol{\epsilon}_g$ carries a smaller guidance coefficient than ours.
The deployed modulation, by contrast, uses the base flow directly, $\hat{\mathbf{v}}_0 + s\gamma\,\hat{\mathbf{g}}$ with $a = \cos(\hat{\mathbf{v}}_0, \hat{\mathbf{g}})$, whereas the exact analogue under the conditional anchor is $\hat{\mathbf{v}}_{\text{cond}} + (s\gamma - 1)\,\hat{\mathbf{g}}$ with $a = \cos(\hat{\mathbf{v}}_{\text{cond}} - \hat{\mathbf{g}},\, \hat{\mathbf{g}})$, so its gated base is obtained by subtraction.

We train the conditional-anchor student with the identical rank, curriculum, teacher, optimizer, and budget, using the composed loss $\|\hat{\mathbf{v}}_{\text{cond}} + (s-1)\,\hat{\mathbf{g}} - \mathbf{v}_{\text{cfg}}^{\text{teacher}}\|^2$.
Table~\ref{tab:anchor_core} first removes modulation and warping to isolate the parameterization, and Table~\ref{tab:anchor_full} then applies the deployed path.

\begin{table}[H]
  \caption{Anchor ablation without modulation or warping (Core), FID-50K ($\downarrow$) on shared class-balanced seed lists.}
  \label{tab:anchor_core}
  \centering
  \small
  \begin{tabular}{lccccc}
    \toprule
    Anchor & $(25,1)$ & $(25,2)$ & $(8,4)$ & $(25,8)$ & $(4,8)$ \\
    \midrule
    Unconditional, $\mathbf{v}_0 + s\,\mathbf{g}$ (ours) & 9.92 & 3.14 & 11.48 & 23.65 & 73.41 \\
    Conditional, $\mathbf{v}_{\text{cond}} + (s-1)\,\mathbf{g}$ & 9.84 & 3.18 & 11.44 & 23.52 & 72.80 \\
    \bottomrule
  \end{tabular}
\end{table}

At the Core level the per-cell gaps of 0.04 to 0.61 FID are of the same order as the per-seed standard deviations of \method-Full on these cells (0.07 to 0.51, Table~\ref{tab:fid_stds}), so we read the two anchors as matching within seed variation.
The conditional anchor is marginally lower at $s{=}1$, where its first branch targets $\mathbf{v}_{\text{cond}}$ directly, and at high $s$, consistent with its smaller guidance coefficient.
This parity is what the exact equivalence of the two forms predicts.

\begin{table}[H]
  \caption{Anchor ablation under the deployed modulation and warping path, FID-50K ($\downarrow$).}
  \label{tab:anchor_full}
  \centering
  \small
  \begin{tabular}{lcc}
    \toprule
    Anchor & $(25,8)$ Full & $(4,8)$ Full \\
    \midrule
    Unconditional (ours) & \textbf{19.28} & \textbf{48.95} \\
    Conditional & 20.41 & 51.58 \\
    \bottomrule
  \end{tabular}
\end{table}

Under the deployed path the anchors diverge by 1.1 to 2.6 FID in favor of the unconditional form, because the alignment gate receives the base flow directly in our form and by subtraction in the conditional one.
Component diagnostics on fixed $(\mathbf{x}_t, c, t)$ states agree with this gap. At $t{=}0.5$ our learned $\hat{\mathbf{v}}_0$ reaches cosine 0.968 to the teacher's $\mathbf{v}_0$, while the conditional anchor's derived base $\hat{\mathbf{v}}_{\text{cond}} - \hat{\mathbf{g}}$ reaches 0.941.
A condition-invariance probe also addresses whether the unconditional branch ignores useful conditioning. Fixing $(\mathbf{x}_t, t)$ and varying $c$, our $\hat{\mathbf{v}}_0$ keeps pairwise cosine similarity of at least 0.98 across conditions, so the branch learns a nearly condition-independent base flow rather than absorbing the conditioning signal.
The conditional anchor is therefore a legitimate alternative whose Core quality matches ours within noise, and we keep the unconditional anchor because the deployed modulation uses the base flow directly.

\subsection{Qualitative Comparison and Visualizations}
\label{app:qualitative}

\begin{figure}[H]
  \centering
  \includegraphics[width=0.82\textwidth]{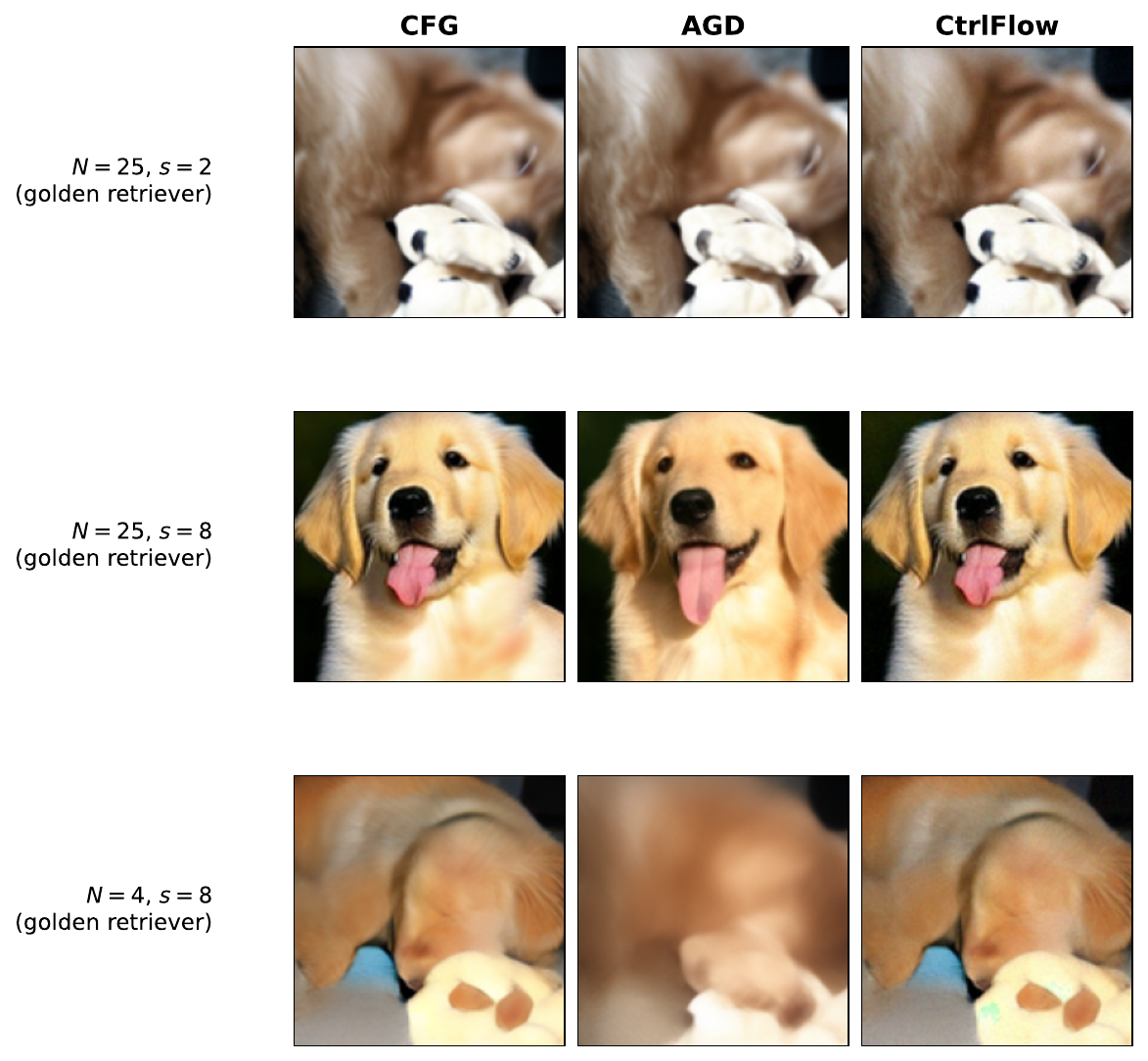}
  \caption{Qualitative comparison at three $(N,s)$ operating points (same ImageNet class, \textit{golden retriever}, per row). Rows show $(N{=}25, s{=}2)$, $(N{=}25, s{=}8)$, and $(N{=}4, s{=}8)$, and columns show CFG ($2N$ passes), the fixed-scale AGD variant ($s{=}2$, $N$ passes, labeled ``AGD''), and \method ($N$ passes). At $(25, 2)$ all methods produce comparable quality. At $(25, 8)$ AGD exhibits color saturation and loss of detail. At $(4, 8)$ AGD collapses to noisy outputs while \method preserves recognizable structure. Aggregate FID values across all 1000 ImageNet classes are reported in Table~\ref{tab:main_results}, and additional classes are in Figure~\ref{fig:extended_qualitative}.}
  \label{fig:qualitative}
\end{figure}

Figure~\ref{fig:qualitative} presents generated samples from CFG, the fixed-scale AGD variant, and \method at three operating points.
At $(25, 2)$, all methods produce visually similar outputs, consistent with the small FID gaps in Table~\ref{tab:main_results}.
At $(25, 8)$, the fixed-scale variant's samples suffer from oversaturation and loss of fine detail, reflecting its Recall collapse to 0.14.
\method samples remain visually comparable to CFG at this scale.
The contrast is most pronounced at $(4, 8)$, where the fixed-scale variant degenerates to noisy outputs (FID 82.17) while \method preserves recognizable structure (FID 48.95).

\begin{figure}[H]
  \centering
  \includegraphics[width=0.67\textwidth]{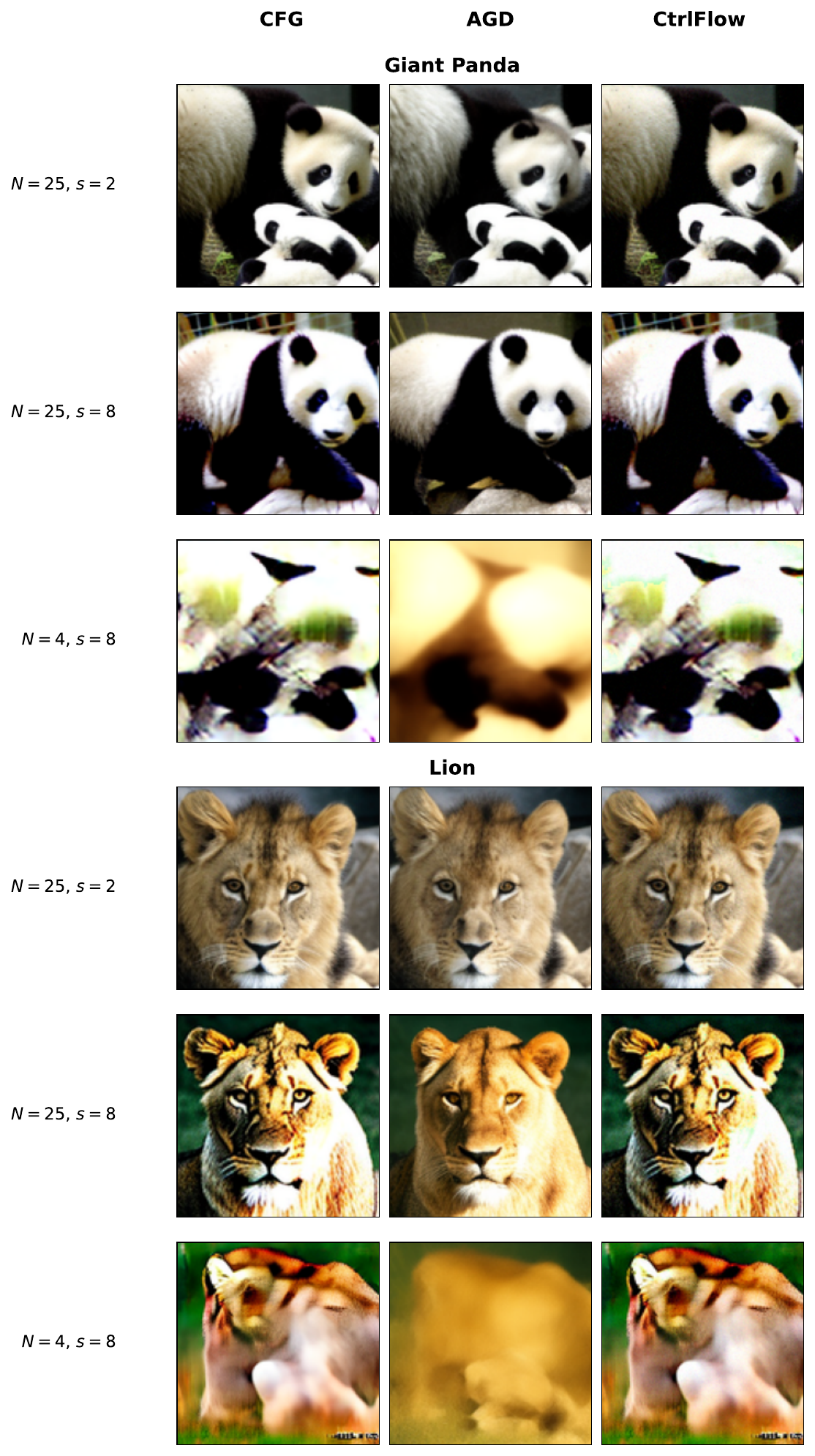}
  \caption{Extended qualitative comparison on two additional ImageNet classes (\textit{giant panda}, \textit{lion}) across the same three $(N, s)$ operating points and three methods as Figure~\ref{fig:qualitative}. Each block of three rows corresponds to one class, and columns show CFG ($2N$ passes), the fixed-scale AGD variant ($N$ passes, labeled ``AGD''), and \method ($N$ passes) at a shared seed. The visual story is consistent across classes. AGD oversaturates and collapses at large $s$, \method tracks CFG closely, and the severity tracks the FID gaps reported in Table~\ref{tab:main_results}.}
  \label{fig:extended_qualitative}
\end{figure}

\subsection{Extended Interpretability Maps}
\label{app:interpretability_extended}

\begin{figure}[H]
  \centering

  \medskip
  \textbf{(a) Butterfly}\\[4pt]
  \includegraphics[width=0.82\textwidth]{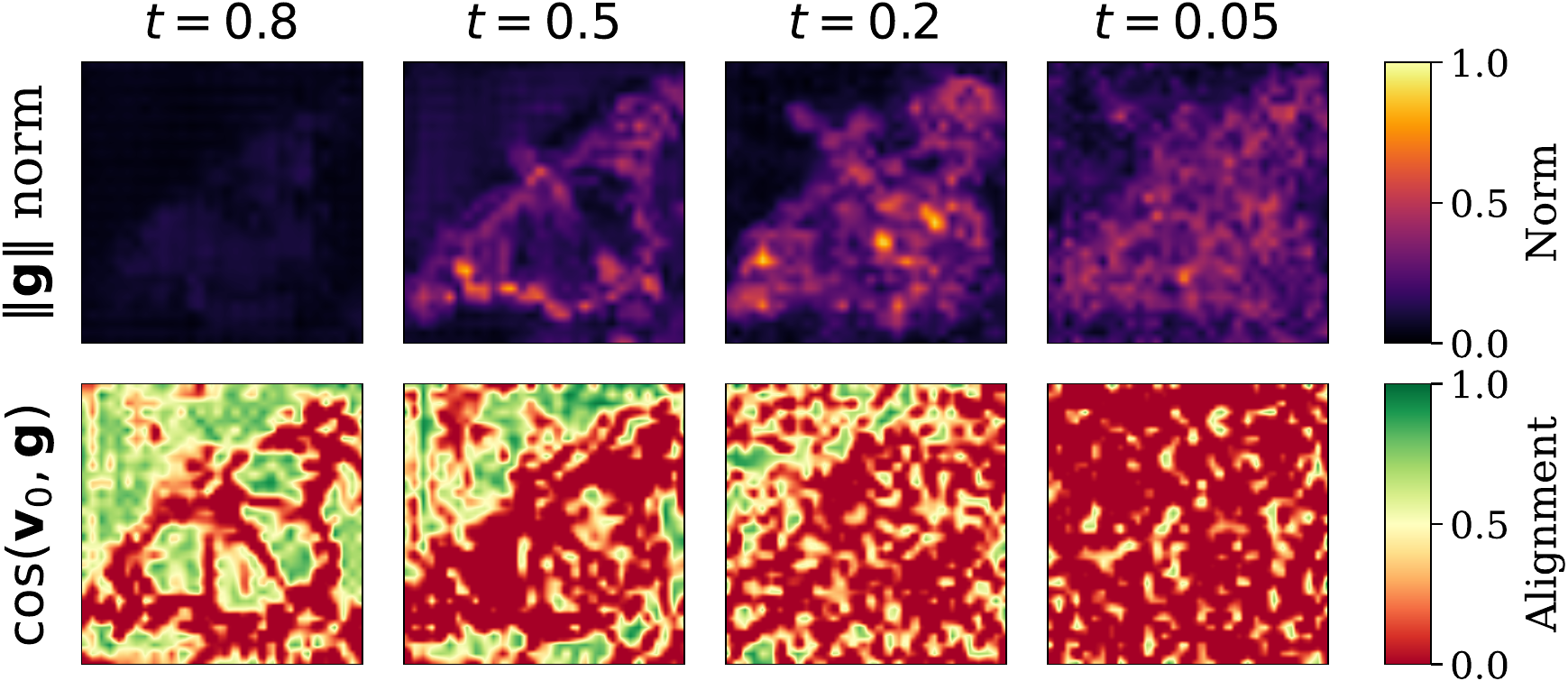}

  \bigskip
  \textbf{(b) Volcano}\\[4pt]
  \includegraphics[width=0.82\textwidth]{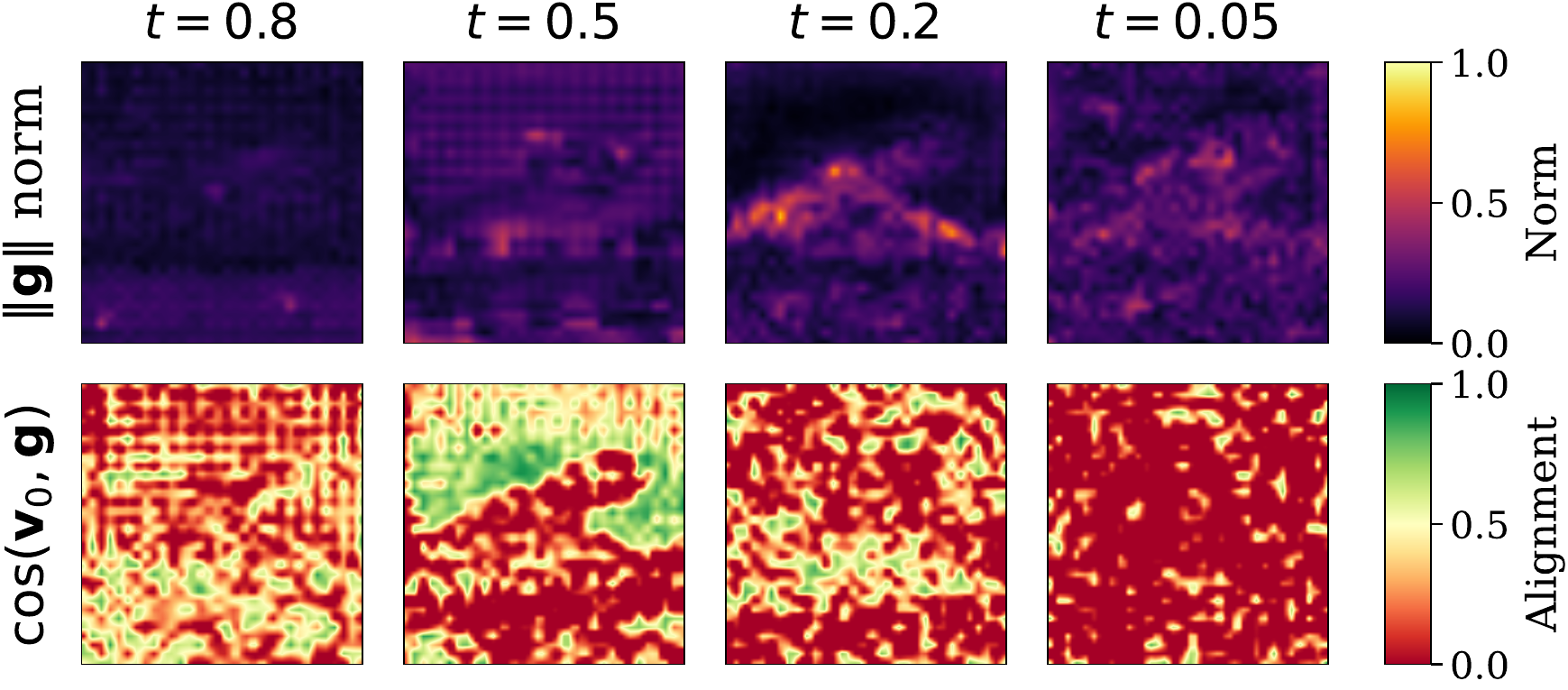}

  \caption{Cross-class interpretability maps (extending Figure~\ref{fig:interpretability}, which shows \textit{golden retriever}). Each panel shows guidance norm $\|\mathbf{g}\|$ (top row) and alignment $\cos(\mathbf{v}_0, \mathbf{g})$ (bottom row) at four timesteps. The temporal pattern matches the main-text figure. Guidance is diffuse at $t{=}0.8$ and concentrates on class-discriminative regions by $t{=}0.05$. The spatial focus varies by class, with wing patterns for \textit{butterfly} and crater and smoke for \textit{volcano}, confirming that $\mathbf{g}$ encodes class-specific semantic structure.}
  \label{fig:interpretability_extended}
\end{figure}

Figure~\ref{fig:interpretability} in the main text shows the guidance norm and alignment maps for \textit{golden retriever}.
Figure~\ref{fig:interpretability_extended} extends this analysis to two additional classes, \textit{butterfly} and \textit{volcano}, to test whether the observed patterns generalize.

The temporal progression is consistent across both additional classes.
At $t{=}0.8$, guidance norms are spatially diffuse regardless of class identity. The model has not yet resolved object structure at this stage, so conditioning acts as a global bias.
By $t{=}0.2$ to $t{=}0.05$, guidance concentrates on class-discriminative regions. For \textit{butterfly}, these are wing patterns and body contours. For \textit{volcano}, they are the crater rim and smoke plume.
Together with the golden retriever maps in the main text, where guidance focuses on facial features and fur texture, the class-specific spatial concentration confirms that the learned $\mathbf{g}$ encodes semantically meaningful conditioning adjustments rather than a generic sharpening signal.

The alignment maps reveal the same orthogonal-to-aligned progression in both classes.
At early timesteps, $\cos(\mathbf{v}_0, \mathbf{g}) \approx 0$, and as the trajectory converges, alignment increases toward $1$.
The consistency of this pattern across three visually distinct categories (fine-grained texture, object with background, scene-level structure) supports the claim that the decomposition captures a general structural property of the guided velocity field.

\subsection{Quantitative Localization of Guidance Maps}
\label{app:localization}

The maps above are qualitative and cover few classes.
To test whether guidance concentrates on the conditioned object, we compute guidance-norm maps on forward-noised ImageNet validation images and score them against the ground-truth object boxes.
We report pointing accuracy~\citep{zhang2016excitation}, the fraction of images whose map maximum falls inside a box, energy-in-box, the fraction of map mass inside the boxes, and deletion and insertion AUC~\citep{petsiuk2018rise}, which track how classifier confidence changes as the highest-scoring pixels are removed from or added to the image.
All four scores are oriented so that higher is better.
Grad-CAM~\citep{selvaraju2017gradcam} maps and a uniformly random map serve as references.

\begin{table}[H]
  \caption{Localization of guidance-norm maps against ground-truth boxes ($\uparrow$ for all columns).}
  \label{tab:localization}
  \centering
  \small
  \begin{tabular}{lcccc}
    \toprule
    Map & Pointing acc. & Energy-in-box & Deletion AUC & Insertion AUC \\
    \midrule
    Guidance norm, \method & \textbf{0.81} & \textbf{0.613} & \textbf{0.624} & \textbf{0.681} \\
    Guidance norm, two-pass teacher & 0.80 & 0.601 & 0.612 & 0.669 \\
    Grad-CAM & 0.74 & 0.548 & 0.587 & 0.642 \\
    Random & 0.14 & 0.219 & 0.501 & 0.503 \\
    \bottomrule
  \end{tabular}
\end{table}

The learned maps localize the object better than Grad-CAM on every score and closely track the maps computed from the two-pass teacher, which grounds the interpretability claim of Section~\ref{subsec:interpretability} quantitatively.

\subsection{Modulation and Warping Visualization}
\label{app:modulation_profiles}

\begin{figure}[H]
  \centering
  \includegraphics[width=0.87\linewidth]{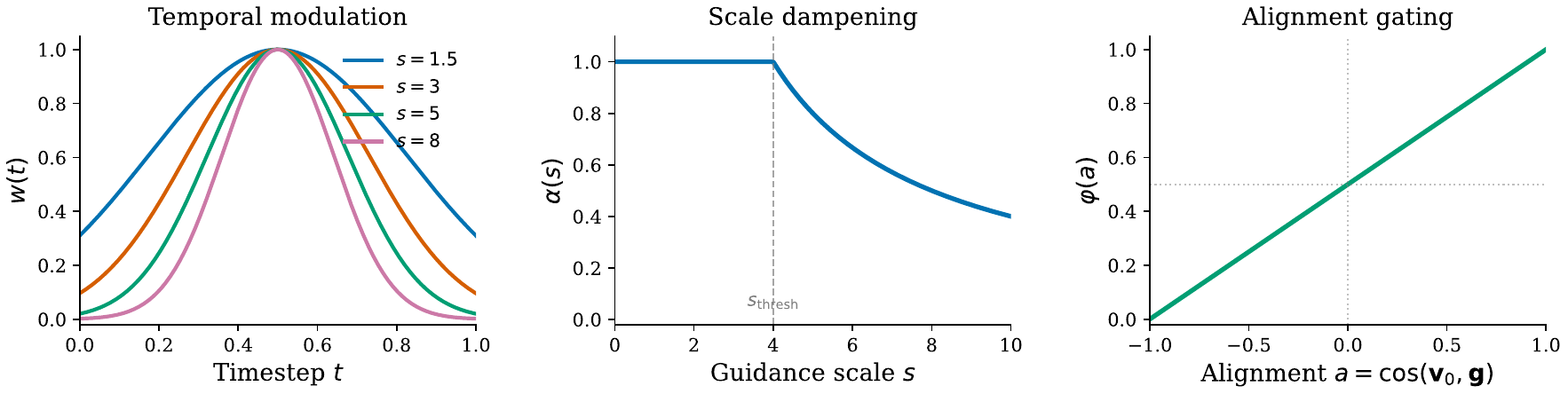}
  \caption{Guidance modulation profiles. \textbf{Left.} Temporal modulation $w(t)$ narrows with increasing $s$. \textbf{Center.} Scale dampening $\alpha(s)$ activates above $s_{\text{thresh}}$. \textbf{Right.} Alignment gating $\varphi(a)$ attenuates guidance under misalignment.}
  \label{fig:modulation_profiles}
\end{figure}

Figure~\ref{fig:modulation_profiles} illustrates the three guidance modulation functions.
The temporal window $w(t)$ narrows as $s$ increases, concentrating guidance in the mid-timestep range.
The dampening function $\alpha(s)$ engages only beyond $s_{\text{thresh}} = 4$, leaving low-scale generation unmodified.
The alignment gate $\varphi(a)$ attenuates guidance when $\mathbf{v}_0$ and $\mathbf{g}$ are misaligned.

\subsection{Time-Warp Visualization}
\label{app:time_warp_vis}

\begin{figure}[H]
  \centering
  \includegraphics[width=0.73\linewidth]{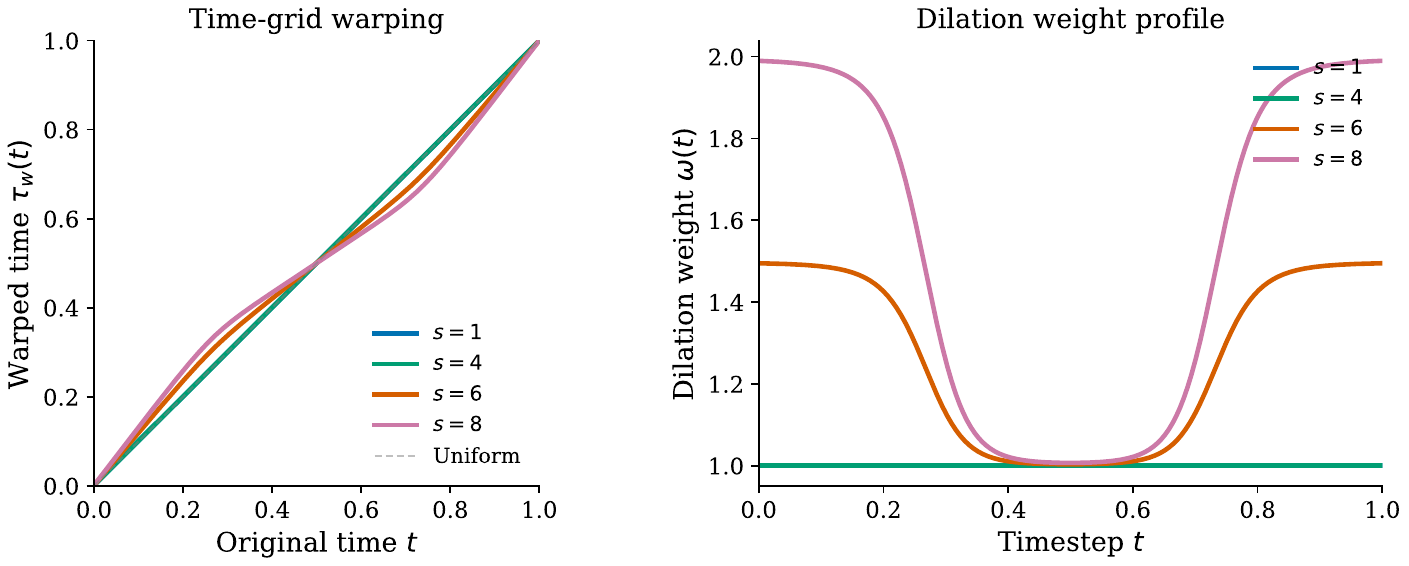}
  \caption{Time-grid warping. \textbf{Left.} Warped time $\tau(t)$ deviates from the uniform diagonal at high $s$. \textbf{Right.} Dilation weight $\omega(t)$ increases at trajectory endpoints for large guidance scales.}
  \label{fig:time_warp}
\end{figure}

Figure~\ref{fig:time_warp} visualizes the time-grid warping curves at different guidance scales.
At $s{=}1$, the schedule remains uniform.
At $s{=}8$, the dilation weight reaches $1.5\times$ at the trajectory endpoints, concentrating solver evaluations in the mid-timestep region where the guided velocity changes most rapidly.

\subsection{FID vs.\ NFE}
\label{app:fid_vs_nfe}

\begin{figure}[H]
  \centering
  \includegraphics[width=0.55\linewidth]{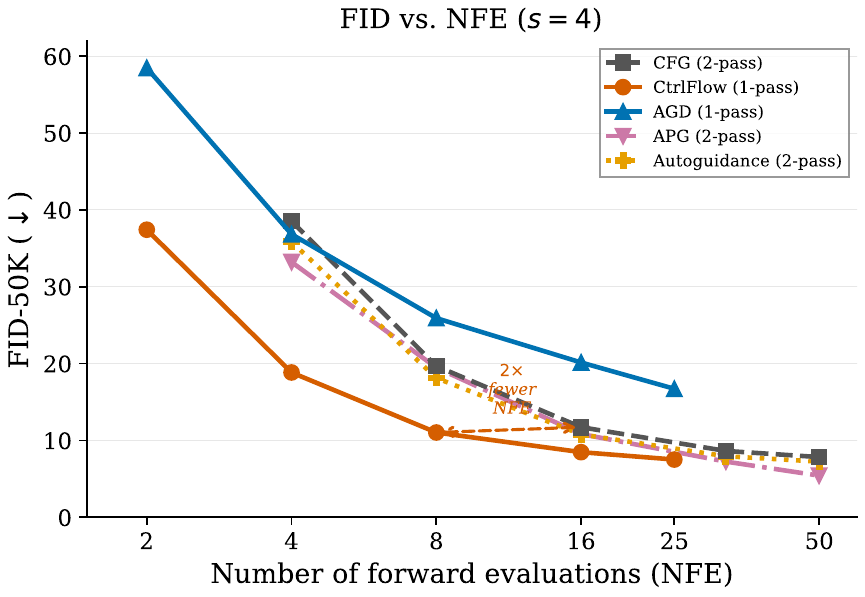}
  \caption{FID-50K vs.\ NFE at $s{=}4$. \method (solid orange) at NFE${}=N$ achieves comparable FID to CFG (dashed gray) at NFE${}=2N$, halving compute for equivalent quality.}
  \label{fig:fid_vs_nfe}
\end{figure}

Figure~\ref{fig:fid_vs_nfe} plots FID against NFE at $s{=}4$, confirming that \method at NFE${}=N$ matches CFG at NFE${}=2N$.

\section{Reproducibility}
\label{app:reproducibility}

This appendix specifies the baseline protocol, computational cost breakdown, and implementation details required to reproduce every result in Section~\ref{sec:experiments}.

\subsection{Baseline Protocol}
\label{app:baseline_protocol}

Table~\ref{tab:baseline_protocol} records the code source, training configuration, and evaluation scope for each baseline.
Every two-pass baseline runs its official public implementation with default hyperparameters on DiT-XL/2 ImageNet 256$\times$256.
Both AGD variants are trained by us, as described in the protocol note below, and GFT is evaluated from its released checkpoint.
We tuned \method's modulation and warping hyperparameters on a 5K-image validation set described in Appendix~\ref{app:hyperparams}.

\begin{table}[H]
  \caption{Baseline protocol. All methods use DiT-XL/2 on ImageNet 256$\times$256. ``Train $s$'' is the guidance scale used during training. ``Eval $s$'' is the range tested. ``$s$-free'' indicates whether one checkpoint supports variable $s$ at inference without retraining.}
  \label{tab:baseline_protocol}
  \centering
  \scriptsize
  \setlength{\tabcolsep}{3pt}
  \begin{tabular}{lllcccc}
    \toprule
    Method & Source & Hyperparams & Train $s$ & Eval $s$ & Models & $s$-free \\
    \midrule
    CFG & Official DiT & Default & N/A & $[1,12]$ & 1 & $\checkmark$ \\
    CFG-Zero* & Official repo & Default & N/A & $[1,8]$ & 1 & $\checkmark$ \\
    APG & Official repo & Default & N/A & $[1,8]$ & 1 & $\checkmark$ \\
    Autoguidance & Official repo & Default & N/A & $[1,8]$ & 1 & $\checkmark$ \\
    CFG + Interval & Official repo & Default & N/A & $[1,8]$ & 1 & $\checkmark$ \\
    AGD (fixed $s$) & Ours, AGD design & Ablation & 2.0 & $[1,8]$ & 1 & $\times$ \\
    AGD-MultiScale & Ours, AGD design & Matched & $\mathcal{U}[1,8]$ & $[1,8]$ & 1 & $\checkmark$ \\
    GFT & Released ckpt. & Default & $\beta_{\text{GFT}} \in [0,1]$ & $[1,8]$ & 1 & $\checkmark$ \\
    \textbf{\method} & Ours & Tuned (App.~\ref{app:hyperparams}) & $\mathcal{U}[1,8]$ & $[1,12]$ & 1 & $\checkmark$ \\
    \bottomrule
  \end{tabular}
\end{table}

\paragraph{AGD Protocol Note.}
As published, AGD~\citep{sadat2025agd} is scale-conditioned. Its adapter receives the guidance scale as an embedded network input and is trained over a sampled range of scales ($\mathcal{U}[1, 6]$ for DiT-XL/2), so one checkpoint serves many scales.
No DiT-XL/2 AGD checkpoint is publicly available, so we train two variants on our frozen teacher and trajectories.
AGD-MultiScale follows the published design, with the scale range widened to $\mathcal{U}[1, 8]$ to match our training range, rank 128 (34M parameters) to match \method's capacity, and the same curriculum and optimizer.
It is the faithful AGD baseline and the primary single-pass comparison.
AGD (fixed $s{=}2$) is an ablation of that design trained at a single scale (14M parameters). We evaluate it at other scales to measure single-checkpoint behavior outside the training scale, which is not how AGD is intended to be used, and its high-scale failure should be read in that light.
The submitted version of this paper described this row as AGD's public $s{=}2$ checkpoint. It is our own single-scale training of the AGD design, and this version corrects the description.

\subsection{Computational Cost}
\label{app:cost}

\begin{table}[H]
  \caption{Computational cost comparison on DiT-XL/2 at $N{=}25$.}
  \label{tab:cost}
  \centering
  \small
  \begin{tabular}{lccccc}
    \toprule
    Method & Params & Extra & NFE & ms/img & Train (GPU-hr) \\
    \midrule
    CFG & 675M & 0\% & $2N$ & 843 & 0 \\
    CFG-Zero* & 675M & 0\% & $2N$ & 854 & 0 \\
    APG & 675M & 0\% & $2N$ & 852 & 0 \\
    Autoguidance & 1015M & 50\% & $2N$ & 891 & 120 \\
    \midrule
    AGD (fixed $s$) & 689M & 2\% & $N$ & 432 & 32 \\
    \textbf{\method} & \textbf{709M} & \textbf{5\%} & $\boldsymbol{N}$ & \textbf{437} & \textbf{48} \\
    \bottomrule
  \end{tabular}
\end{table}

\paragraph{Latency Breakdown.}
The two branches share one frozen backbone pass per step, so backbone compute is incurred once and $\text{NFE} = N$ is exact.
The speedup is $1.93\times$ rather than $2\times$ because half of CFG's 843\,ms, two sequential evaluations per step, is 421.5\,ms, and the adapters and modulation add 15.5\,ms at $N{=}25$ (Table~\ref{tab:latency}).
The single-adapter AGD baseline (Table~\ref{tab:cost}) measures 432\,ms under the same protocol, so the second branch and the gates account for the remaining 5\,ms.

\begin{table}[H]
  \caption{Latency breakdown of \method at $N{=}25$, batch size 1, A100, 50 warmup and 200 timed iterations under CUDA synchronization.}
  \label{tab:latency}
  \centering
  \small
  \begin{tabular}{lc}
    \toprule
    Component & ms/image \\
    \midrule
    Frozen backbone, 25 steps at 16.9\,ms & 421.5 \\
    Adapter matrix multiplications, both branches & 12.6 \\
    Modulation and composition & 2.9 \\
    Warp precomputation (cached) & $<1$ \\
    \midrule
    \textbf{Measured total} & \textbf{437} \\
    \bottomrule
  \end{tabular}
\end{table}

\subsection{Implementation Details}
\label{app:implementation_details}

\paragraph{LoRA Placement.}
Both adapters attach rank-64 LoRA modules to the query, key, value, and output projections of every self-attention layer in the DiT backbone.
MLP layers and layer normalization parameters remain frozen.
The two branches do not keep separate hidden states. The frozen backbone computes one hidden-state stream per step, each LoRA module reads that shared stream, and each branch's low-rank residuals are accumulated into its own velocity-shaped output rather than written back into the stream.
The two outputs are combined as $\hat{\mathbf{v}}_0 + s \cdot \hat{\mathbf{g}}$ after the forward pass.

\paragraph{Teacher Target Construction.}
For each training sample $(\mathbf{x}_t, c, t, s)$, the frozen teacher performs two forward passes, one unconditional ($c = \varnothing$) and one conditional ($c$).
The resulting teacher target is $\mathbf{v}_{\text{cfg}}^{\text{teacher}} = \mathbf{v}_\theta(\mathbf{x}_t, \varnothing, t) + s \cdot [\mathbf{v}_\theta(\mathbf{x}_t, c, t) - \mathbf{v}_\theta(\mathbf{x}_t, \varnothing, t)]$.
Teacher outputs are generated online rather than cached.

\paragraph{Solver and Evaluation.}
Inference relies on DPM-Solver-2 (second-order, multistep) with velocity prediction and the uniform-then-warped timestep schedule.
FID-50K is computed through the ADM evaluation suite using the standard ImageNet-256 Inception statistics file.
All methods share the same three 50K class-balanced generation seed lists.

\paragraph{Precision and Hardware.}
Training and inference run in bf16 mixed precision with TF32 enabled on 8$\times$A100 (80\,GB).
The AGD-MultiScale baseline and 3-seed evaluation runs were executed on 4$\times$H200 (141\,GB) via Modal cloud compute.
All methods share the same ADM evaluation pipeline, class-balanced seed lists, and precision settings.
Small hardware-dependent numerical differences may remain across GPU architectures.
Timing measurements use the A100 setup with batch size 1, 50 warmup iterations, and 200 timed iterations under CUDA synchronization.

\paragraph{Validation Protocol.}
Modulation and warping hyperparameters were tuned on a held-out 5K-image validation set spanning five $(N, s)$ configurations, namely $(25, 2)$, $(25, 8)$, $(8, 4)$, $(4, 2)$, and $(4, 8)$.
These five configurations overlap with the headline test configurations.
We acknowledge this overlap as a limitation.
Two checks use configurations never seen during selection. \method matches CFG on all 20 untuned grid configurations except $(2, 8)$, and the mechanism-transfer test of Appendix~\ref{app:mechanism_transfer} includes the untuned $(16, 6)$ and $(8, 6)$.
The ablation tables in Appendices~\ref{app:ablation_component} to~\ref{app:rank_ablation} provide further robustness evidence across hyperparameter variations.

\paragraph{Additional Evaluations.}
Class adherence uses a frozen torchvision ResNet-50 (IMAGENET1K\_V2) on the FID seed lists (Appendix~\ref{app:class_adherence}).
The GFT comparison uses the released DiT-XL/2 ImageNet-256 checkpoint with its own sampler at matched NFE (Appendix~\ref{app:gft}).
The text-to-image pilot uses Stable Diffusion 2.1 with zero-shot COCO FID-10K and CLIPScore (Appendix~\ref{app:sd21}).

\clearpage
\section*{NeurIPS Paper Checklist}

\begin{enumerate}

\item {\bf Claims}
    \item[] Question: Do the main claims made in the abstract and introduction accurately reflect the paper's contributions and scope?
    \item[] Answer: \answerYes{}
    \item[] Justification: Main claims in the abstract and introduction (96\% CFG-match accuracy against 20 of 25 for a matched scale-conditioned opaque student, $1.93\times$ speedup, interpretable decomposition) are supported by experiments in Section~\ref{sec:experiments}, ablations in Section~\ref{subsec:ablations}, and the matched-baseline, mechanism-transfer, class-adherence, and text-to-image results in Appendix~\ref{app:extended}.

\item {\bf Limitations}
    \item[] Question: Does the paper discuss the limitations of the work performed by the authors?
    \item[] Answer: \answerYes{}
    \item[] Justification: Section~\ref{sec:conclusion} includes an explicit Limitations paragraph discussing two-backbone evaluation (DiT-XL/2 and DiT-L/4), text-to-image evidence limited to a single-configuration Stable Diffusion 2.1 pilot, untested video and higher resolutions, the overlap between tuning and headline configurations, and the deliberate departure from CFG dynamics at high guidance scales.

\item {\bf Theory assumptions and proofs}
    \item[] Question: For each theoretical result, does the paper provide the full set of assumptions and a complete (and correct) proof?
    \item[] Answer: \answerYes{}
    \item[] Justification: Appendices~\ref{app:gradient},~\ref{app:error}, and~\ref{app:timewarp} provide supporting analysis with stated assumptions. Formal results include gradient scaling under the unified loss (Claims~\ref{prop:conflict} to~\ref{prop:scaling}), decomposition identifiability under scale variation (Proposition~\ref{prop:identifiability}), per-step velocity error bounds (Propositions~\ref{prop:perstep} and~\ref{prop:trajectory}), the exact bias-error decomposition of the modulated velocity (Proposition~\ref{prop:bias_error}), schedule monotonicity (Proposition~\ref{prop:monotone}), and curvature-weighted step allocation (Proposition~\ref{prop:truncation}). All assumptions are stated explicitly. The identifiability result assumes realizability of the adapter function classes. In practice, low-rank adapters provide an approximation, and decomposition fidelity is validated empirically (Section~\ref{subsec:interpretability}). The Jacobian-comparability condition behind the gradient-scaling claim is measured empirically (Appendix~\ref{app:gradient}).

    \item {\bf Experimental result reproducibility}
    \item[] Question: Does the paper fully disclose all the information needed to reproduce the main experimental results of the paper to the extent that it affects the main claims and/or conclusions of the paper (regardless of whether the code and data are provided or not)?
    \item[] Answer: \answerYes{}
    \item[] Justification: Section~\ref{subsec:exp_setup} specifies the backbone, dataset, evaluation protocol, and training details. Appendix~\ref{app:hyperparams} provides all hyperparameter values, and Appendix~\ref{app:reproducibility} specifies the baseline protocol, cost measurement, and implementation details.

\item {\bf Open access to data and code}
    \item[] Question: Does the paper provide open access to the data and code, with sufficient instructions to faithfully reproduce the main experimental results, as described in supplemental material?
    \item[] Answer: \answerNo{}
    \item[] Justification: Code is not included with the paper. ImageNet is publicly available, and the paper and appendix provide sufficient detail for independent reimplementation.

\item {\bf Experimental setting/details}
    \item[] Question: Does the paper specify all the training and test details (e.g., data splits, hyperparameters, how they were chosen, type of optimizer) necessary to understand the results?
    \item[] Answer: \answerYes{}
    \item[] Justification: Section~\ref{subsec:exp_setup} specifies the LoRA rank, curriculum, and training budget, and Appendix~\ref{app:hyperparams} provides the optimizer, learning rates, batch size, phase durations, and the complete hyperparameter table.

\item {\bf Experiment statistical significance}
    \item[] Question: Does the paper report error bars suitably and correctly defined or other appropriate information about the statistical significance of the experiments?
    \item[] Answer: \answerYes{}
    \item[] Justification: All 25 configurations are evaluated over 3 generation seeds. Standard deviations are reported in Appendix~\ref{app:fid_stds} (Table~\ref{tab:fid_stds}). Representative stds range from $\pm$0.07 at easy configs to $\pm$0.94 at the extreme $(2,8)$ failure case. The 24/25 CFG-match classification is stable across all seeds.

\item {\bf Experiments compute resources}
    \item[] Question: For each experiment, does the paper provide sufficient information on the computer resources (type of compute workers, memory, time of execution) needed to reproduce the experiments?
    \item[] Answer: \answerYes{}
    \item[] Justification: Training requires 48 GPU-hours on 8$\times$A100 (80\,GB). Inference timing is reported in Table~\ref{tab:cost}. Total project compute, including preliminary experiments, baselines, evaluation, and the experiments added during review, was approximately 1,400 GPU-hours, part of which ran on Modal cloud compute supported by Modal academic research credits.
    
\item {\bf Code of ethics}
    \item[] Question: Does the research conducted in the paper conform, in every respect, with the NeurIPS Code of Ethics \url{https://neurips.cc/public/EthicsGuidelines}?
    \item[] Answer: \answerYes{}
    \item[] Justification: This work conforms to the NeurIPS Code of Ethics. It proposes an efficiency improvement for existing diffusion models and does not introduce new risks beyond those inherent to image generation.

\item {\bf Broader impacts}
    \item[] Question: Does the paper discuss both potential positive societal impacts and negative societal impacts of the work performed?
    \item[] Answer: \answerYes{}
    \item[] Justification: See the Broader Impacts paragraph in Section~\ref{sec:conclusion}. The method reduces compute cost for guided diffusion but inherits the same misuse risks as any image generation technology.
    
\item {\bf Safeguards}
    \item[] Question: Does the paper describe safeguards that have been put in place for responsible release of data or models that have a high risk for misuse (e.g., pre-trained language models, image generators, or scraped datasets)?
    \item[] Answer: \answerNA{}
    \item[] Justification: This paper proposes an inference-efficiency method and does not release a pretrained model or dataset. Safeguards are the responsibility of the base model provider.

\item {\bf Licenses for existing assets}
    \item[] Question: Are the creators or original owners of assets (e.g., code, data, models), used in the paper, properly credited and are the license and terms of use explicitly mentioned and properly respected?
    \item[] Answer: \answerYes{}
    \item[] Justification: DiT-XL/2 is released under a permissive license. ImageNet is used under its standard research license. Stable Diffusion 2.1 is used under the CreativeML Open RAIL++-M license, COCO under its published terms of use, and the torchvision ResNet-50 weights and the released GFT checkpoint under their respective licenses. All baseline implementations and checkpoints are cited with their original papers.

\item {\bf New assets}
    \item[] Question: Are new assets introduced in the paper well documented and is the documentation provided alongside the assets?
    \item[] Answer: \answerNA{}
    \item[] Justification: No new datasets or pretrained models are released with this paper.

\item {\bf Crowdsourcing and research with human subjects}
    \item[] Question: For crowdsourcing experiments and research with human subjects, does the paper include the full text of instructions given to participants and screenshots, if applicable, as well as details about compensation (if any)? 
    \item[] Answer: \answerNA{}
    \item[] Justification: This work does not involve crowdsourcing or research with human subjects.

\item {\bf Institutional review board (IRB) approvals or equivalent for research with human subjects}
    \item[] Question: Does the paper describe potential risks incurred by study participants, whether such risks were disclosed to the subjects, and whether Institutional Review Board (IRB) approvals (or an equivalent approval/review based on the requirements of your country or institution) were obtained?
    \item[] Answer: \answerNA{}
    \item[] Justification: This work does not involve human subjects research.

\item {\bf Declaration of LLM usage}
    \item[] Question: Does the paper describe the usage of LLMs if it is an important, original, or non-standard component of the core methods in this research? Note that if the LLM is used only for writing, editing, or formatting purposes and does \emph{not} impact the core methodology, scientific rigor, or originality of the research, declaration is not required.
    \item[] Answer: \answerNA{}
    \item[] Justification: LLMs are not used as a component of the core method. Any use of LLMs for writing assistance does not impact the methodology, scientific rigor, or originality of the research.

\end{enumerate}

\end{document}